\documentclass[journal]{IEEEtran}
\usepackage{amsmath}
\usepackage[caption=false,font=footnotesize]{subfig}
\usepackage{float}
\usepackage{multirow}

\ifCLASSINFOpdf

  \usepackage[pdftex]{graphicx}
  \usepackage[justification=centering]{caption}
\else
\fi
\usepackage{stfloats}
\begin{document}
%
\title{Attention-SLAM: A Visual Monocular SLAM Learning from Human Gaze }
%
%
%

\author{Jinquan Li,~\IEEEmembership{}
        Ling Pei\IEEEauthorrefmark{1},~\IEEEmembership{Senior Member,~IEEE,}
        Danping Zou,~\IEEEmembership{}
        Songpengcheng Xia,~\IEEEmembership{}
        Qi Wu,~\IEEEmembership{}
        Tao Li,~\IEEEmembership{}
        Zhen Sun,~\IEEEmembership{}
        
        Wenxian Yu,~\IEEEmembership{Senior Member,~IEEE}
\thanks{All authors are associated with Shanghai Jiao Tong University, Shanghai, China.
	    L. Pei is the corresponding author. Emails: ling.pei@sjtu.edu.cn.}
\thanks{}
\thanks{}}

\maketitle

\begin{abstract}
This paper proposes a novel simultaneous localization and mapping (SLAM) approach, namely Attention-SLAM, which simulates human navigation mode by combining a visual saliency model (SalNavNet) with traditional monocular visual SLAM. Most SLAM methods treat all the features extracted from the images as equal importance during the optimization process. However, the salient feature points in scenes have more significant influence during the human navigation process. Therefore, we first propose a visual saliency model called SalVavNet in which we introduce a correlation module and propose an adaptive Exponential Moving Average (EMA) module. These modules mitigate the center bias to enable the saliency maps generated by SalNavNet to pay more attention to the same salient object. Moreover, the saliency maps simulate the human behavior for the refinement of SLAM results. The feature points extracted from the salient regions have greater importance in optimization process. We add semantic saliency information to the Euroc dataset to generate an open-source saliency SLAM dataset. Comprehensive test results prove that Attention-SLAM outperforms benchmarks such as Direct Sparse Odometry (DSO), ORB-SLAM, and Salient DSO in terms of efficiency, accuracy, and robustness in most test cases.

\end{abstract}


\IEEEpeerreviewmaketitle

\section{Introduction}

\IEEEPARstart{S}{imultaneous} localization and mapping (SLAM) aims to self-localize a robot and estimates a model of its environment of information \cite{1}. It has been widely used in Augmented Reality \cite{26} and plays an essential role in the autonomous flight of Unmanned Aerial Vehicle (UAV) \cite{28}, personal indoor navigation \cite{25}. Visual SLAM is an essential branch in SLAM. Traditional methods \cite{3} usually utilize low-level image features to track, which makes the SLAM system unable to pay attention to the global information of the images. Nowadays, many researchers use semantic constraints which makes SLAM methods have better robustness and accuracy.

\begin{figure*}[htbp]
	\centering
	
	\includegraphics[width=0.8\linewidth]{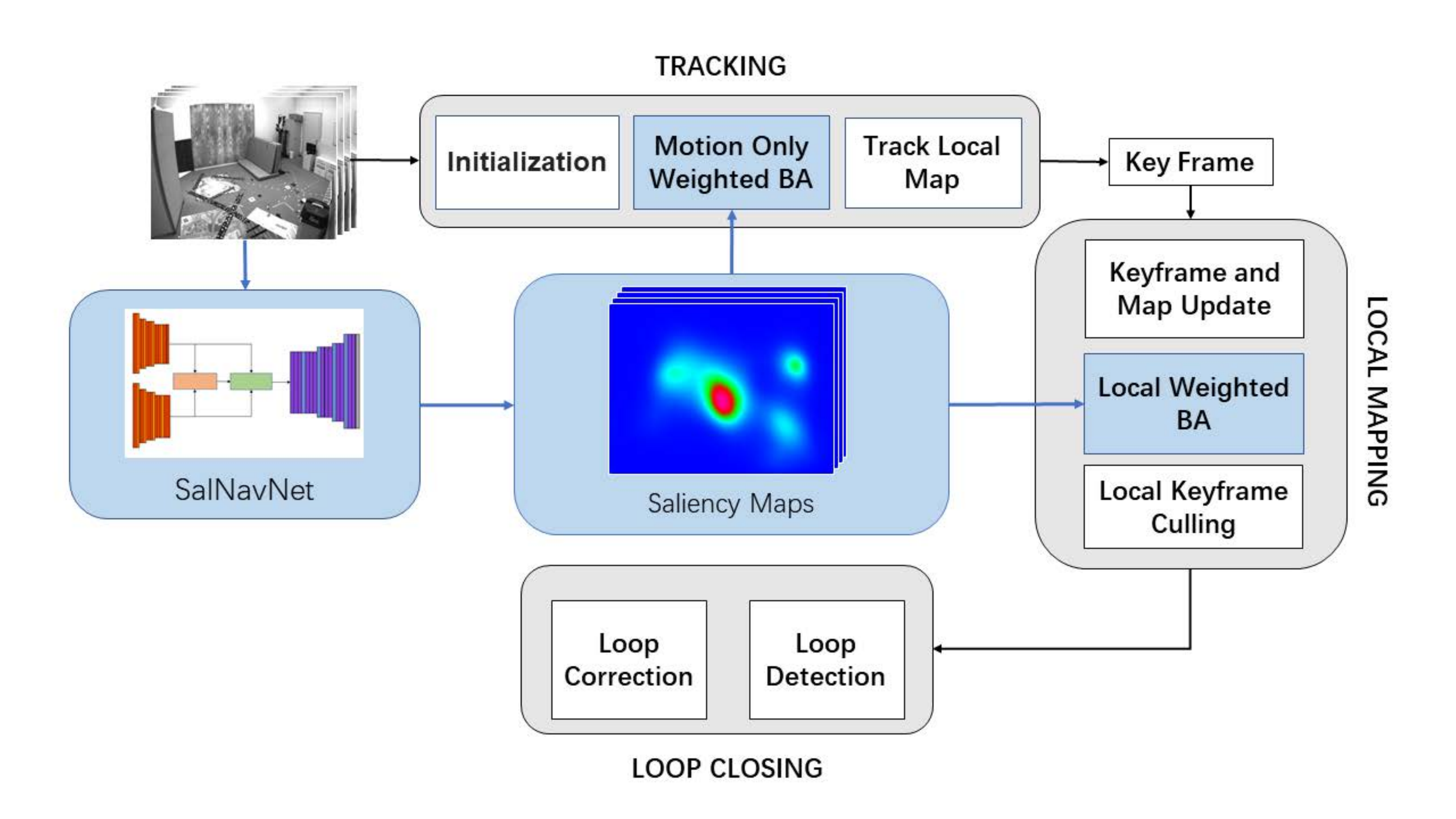}

	\caption{Architecture overview of Attention-SLAM, the blue parts are our contributions.}
	\label{fig: Method Pipeline}
\end{figure*}

In this paper, we focus on adding saliency semantic information to feature-based visual monocular SLAM. A new method is introduced to help this system simulate the behavior pattern of humans when navigating. Humans have the instinct to evaluate and build environmental maps. When people move around in an environment, they usually move eyes to focus on and remember the noticeable landmarks, which usually contain the most valuable semantic information. In order to simulate the navigation pattern of human, we firstly utilize saliency models to generate the corresponding saliency maps of Euroc dataset \cite{9}. These maps show the significant area of each frame in the image sequences. Secondly, we adopt them as weights to make the feature points have different importance in the Bundle Adjustment (BA) process. It helps our system maintain semantic consistency. When there are similar textures in the image sequence, the conventional SLAM method based on the feature points may mismatch. These mismatch points may decrease the accuracy of the SLAM system. Therefore, our approach ensures that the system focuses on the feature points in the most significant areas, which improve accuracy and efficiency. We also utilize information theory to select keyframes and estimate the uncertainty of pose estimation. It proves that our method reduces the uncertainty of motion estimation from the perspective of information theory. The pipeline of our method is shown in Fig. \ref{fig: Method Pipeline}.

The main contributions of this paper are as follows:
\begin{itemize}
	\item We propose a novel SLAM architecture, namely Attention-SLAM. A weighted BA method is proposed to replace the traditional BA methods in SLAM. It can reduce the trajectory error more effectively. Significant features are adopted for the SLAM back-end with high weights by learning from the human gaze during navigating. Compared to the benchmark, Attention-SLAM can reduce the uncertainty of the pose estimation with fewer keyframes and achieve higher accuracy.\\
    \item We propose a visual saliency model called SalNavNet to predict the salient areas in the frames. We introduce a correlation module and propose an adaptive EMA module in SalNavNet. These modules mitigate the center bias of the saliency model and learn the correlation information between frames. By mitigating the center bias that most visual saliency models have, the visual saliency semantic information extracted by SalNavNet can help Attention-SLAM focus on the feature points of the same salient objects consistently.\\
    \item By applying SalNavNet, we generate an open-source saliency dataset based on Euroc \cite{9}. The evaluation using saliency Euroc datasets proves that Attention-SLAM outperforms the benchmark in terms of efficiency, accuracy, and robustness. The dataset is available at https://github.com/Li-Jinquan/Salient-Euroc.

\end{itemize}

\section{Related Work}
\subsubsection{Visual SLAM}
MonoSLAM \cite{12} is the first real-time monocular visual SLAM system. It uses EKF (Extended Kalman Filter) as the back-end, traces sparse features of the front-end, updates the current state of the camera and all the feature points of the state variables. The EKF has been widely used in many SLAM systems \cite{56}\cite{57}.
PTAM \cite{13} is the first keyframe-based SLAM algorithm that separates tracking and mapping as two threads. Many subsequent visual SLAM systems have adopted similar approaches. It is also the first SLAM system that adopts non-linear optimization as a back-end rather than adopts a filter back-end. ORB-SLAM \cite{7} computes around ORB features, including ORB dictionaries for visual odometry and loop detection. The ORB feature is more efficient than SIFT or SURF. ORB-SLAM innovatively uses three threads to complete SLAM: the tracking thread which tracks feature points in real-time, the optimization thread for local Bundle Adjustment, the loop detection and optimization thread which builds global pose graph. Besides, our group uses lines \cite{24} and plane features \cite{27}\cite{29} to enhance the robustness of the SLAM system.  Q. Fu et al. \cite{55} also use line features to improve the trajectory accuracy of the RGB-D SLAM system effectively.

With the development of deep learning, many researchers attempt to combine traditional SLAM methods with deep learning techniques. These approaches extract the semantic information of images to improve the performance of SLAM. Many works show that the combination of semantic information and the SLAM system has promising performances. Mask-SLAM \cite{6} introduces the image segmentation network into the traditional SLAM method. It eliminates the feature points in the background environment and improves the robustness of the algorithm in the dynamic environment. VSO \cite{2} treats semantic information as an invariant of scene representation. Although changes in viewpoint, illumination, and scale affect the low-level representation of objects, these cases do not affect the semantic representation of these objects. VSO also applies a new cost function to minimize semantic reprojection errors and uses semantic constraints to solve the drift problems caused by error accumulation in SLAM during the reconstruction process. Semantic constraints are used to drift corrections for the medium-term continuous tracing of points. N. Brasch et al. \cite{39} add a direct method based on ORB-SLAM, using the measurement information in each frame and the semantic information output by CNN to maintain the map. The maintenance process is completed in a probabilistic form. DynaSLAM \cite{40} utilizes the instance segmentation method to detect moving and possible moving objects. It only adopts the ORB feature points in the static region of the images for pose estimation. Sometimes, when there are few feature points in the static areas, DynaSLAM may not perform well.

As illustrated above, most semantic SLAM methods use semantic segmentation to extract the high-level information of the environment. However, images contain much more information. The significance of images also can provide us more valuable information. Salient DSO \cite{33} combines a saliency model and image segmentation model with Direct Sparse Odometry (DSO) \cite{34}. Zhou et al. \cite{1a} utilize saliency information to extract and match features. This method enhances the performance of DSO. Our method uses state-of-the-art saliency models and proposes a saliency model SalNavNet to extract salient information from images. The semantic saliency information is combined with the sparse-point method, ORB-SLAM, which helps the SLAM system focus on the feature points on the same salient object consistently.
\\
\subsubsection{Visual Saliency Model}
\begin{figure}[hbtp]
	\centering
	
	\includegraphics[width=0.7\linewidth]{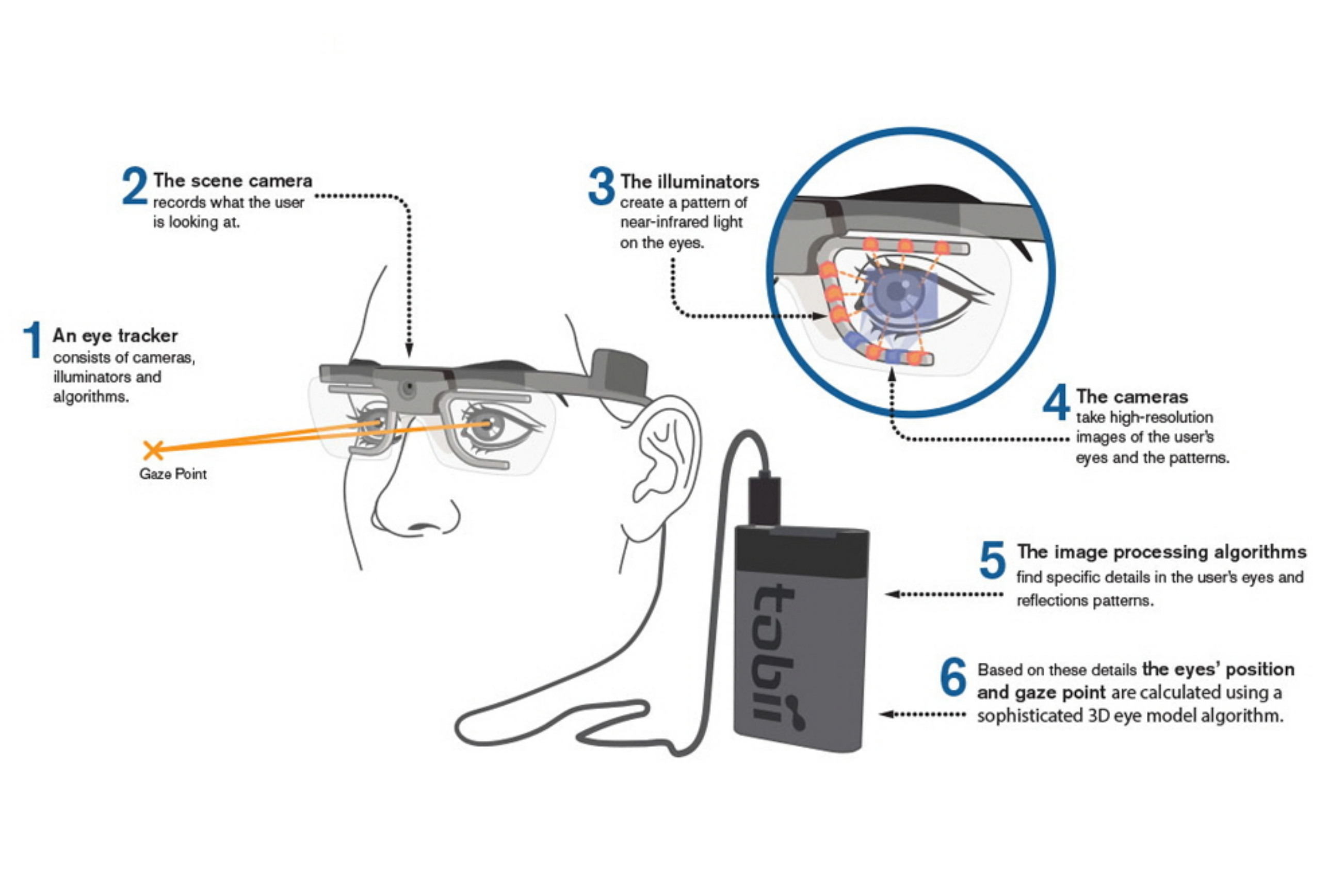}
	\caption{Eye tracking equipment(Tobii), many saliency datasets use this equipment to collect the eye movement data from humans. It can find the areas attracting attention in a environment. }
	\label{fig:eye tracking equipment}
\end{figure}

The attention mechanism can help people filter out critical information quickly. The saliency model has two types \cite{42}: the bottom-up methods (find out crucial information) and the top-down methods (find the corresponding information according to the current task). The bottom-up saliency model is the most extensively studied aspect of visual saliency. By tracking the human gaze, as shown in Fig. \ref{fig:eye tracking equipment}, we can identify the salient parts of the environment. The human gaze can also be utilized to study the significance of pictures and videos.

\begin{figure}[hbtp]
	\centering
	
	\includegraphics[width=0.9\linewidth]{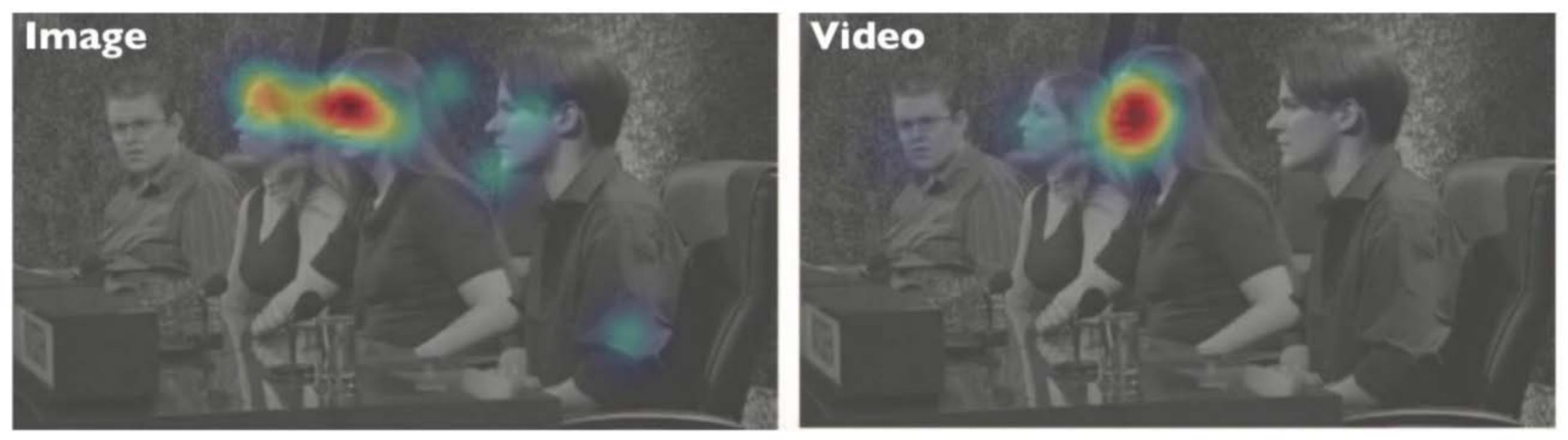}
	\caption{Rudoy et al. \cite{50} show a same image which is observed twice, the left one is a static image mode with 3 seconds gazing while the right one is video mode. }
	\label{fig:Saliency Difference}
\end{figure}

The image saliency model has been thoroughly studied in the past few years, eDN \cite{46} used CNN for the first time to predict the saliency of an image. DeepFix \cite{47} is the first model which applies FCNN for image saliency prediction. Jetley et al. \cite{48} propose a probability distribution prediction, which uses a generalized Bernoulli distribution as a saliency formula and trains the model to learn this distribution. Wang et al. \cite{49} propose a DVA model in which the encoder-decoder architecture is trained on multiple scales to predict pixel-level saliency. Cornia et al. \cite{51} propose SAM-ResNet \& SAM-VGG . It combines a fully convolutional network and a recursive convolutional network, which gives a spatial attention mechanism. Pan et al. \cite{40} ropose SalGAN, which uses Generative Adversarial Network (GAN) to predict saliency. It consists of two modules, a generator and a discriminator. The generator is learned via back-propagation using binary cross-entropy loss on existing saliency maps, which is then passed to the discriminator trained to identify whether the provided saliency map is synthesized by the generator, or built from the actual fixations. 

In recent years, more and more research has been carried out on how human visual attention changed in dynamic scenes. The video saliency model is more common and essential in daily human behavior. Motion information in the video provides reliable guidance for human eye attention detection. Bak et al. \cite{18} propose the classical two-stream network, which combines networks that extract static appearance features with networks that extract motion features. Jiang et al. \cite{19} use a two-layer LSTM network to build a video saliency model. In combination with the network used to detect objectivity, light flow, and static appearances, Wang et al. \cite{10} propose a dynamic saliency model based on the convLSTM \cite{14} network. The model is based on the addition of a static attentive module. The extraction of dynamic and static saliency features is further decoupled. The entire network structure is trained by making full use of existing large-scale static eye movement data. At the same time, the network design avoids the drawbacks of time-consuming optical flow calculation required by previous dynamic saliency models. Linardos et al. \cite{8} introduce ConLSTM to build a SalLSTM model, which achieves excellent results. So we use this model to predict human eye fixations to find out the crucial areas that attract attention in dynamic scenes.

\section{Methodology}
Attention-SLAM consists of two components, the first part is the preprocessing of the input data, and the second part is a visual SLAM system. In the first part, we generate the saliency map corresponding to the SLAM dataset using proposed SalNavNet. These saliency maps are used as inputs to help the SLAM system find the salient key points. The second part of the visual SLAM system uses the saliency map to improve optimization accuracy and efficiency. Each component of Attention-SLAM is discussed in detail. The pipeline of Attention-SLAM is shown in Fig. \ref{fig: Method Pipeline}.

\subsection{SalNavNet}

In the field of visual saliency, the video saliency prediction. Saliency predictions can be defined as finding a stimuli-saliency mapping function:
\begin{equation}
min\sum_{k=1}^{K} \sum_{n=1}^{N} E\left(I_{n}, p_{n}^{k}\right)
\label{equation: eye tracking}
\end{equation}

where $I_{n}$ is the saliency masks of the $n$-th image predicted by saliency model, $p_{n}^{k}$ is the groud-truth data of the $k$-th observer when watching the $n$-th image. $ E\left(I_{n}, p_{n}^{k}\right) $ represents the error between the real value of the human gaze, and the model predicted value in every frame.
\begin{figure*}[h]
	\centering	
	\includegraphics[width=0.9\linewidth]{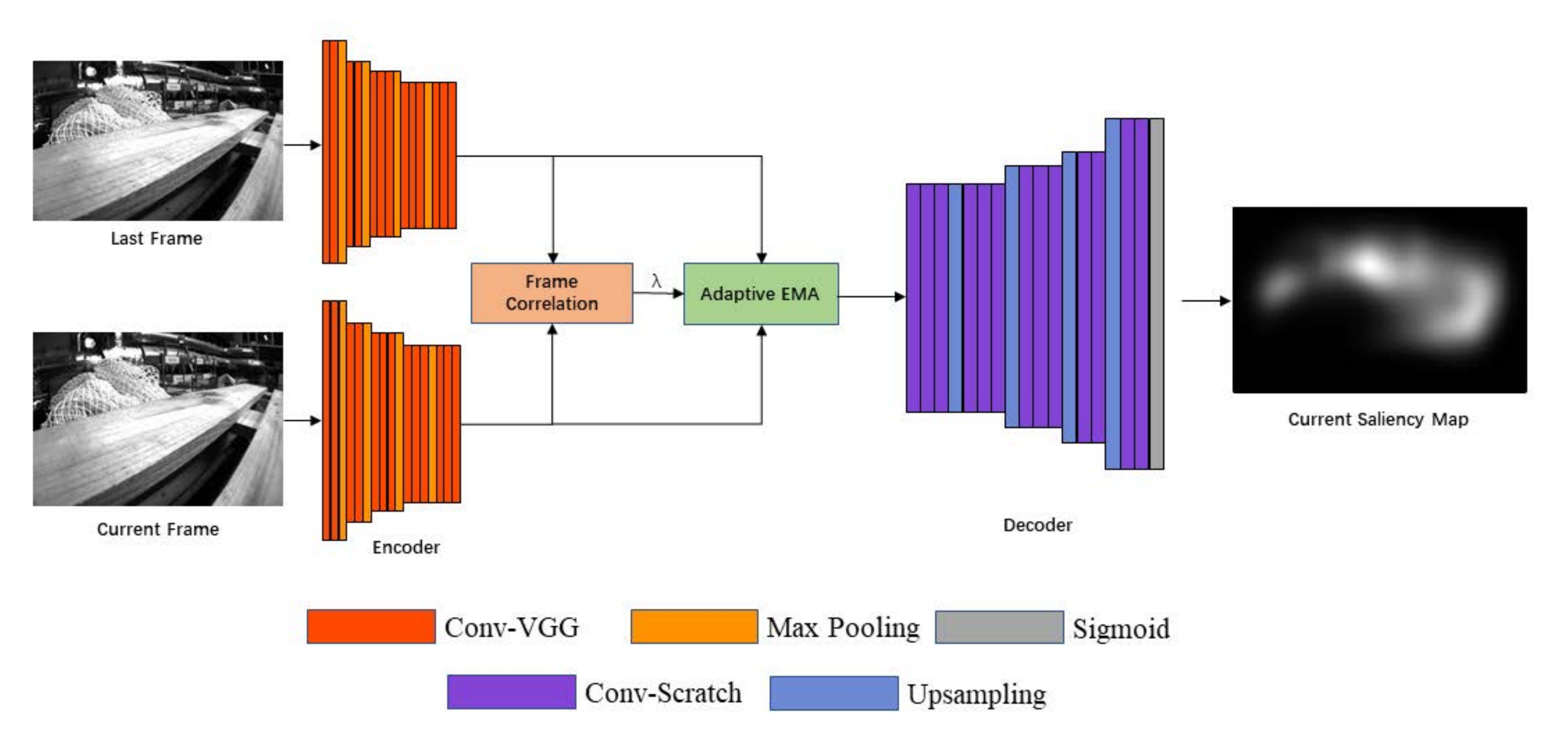}
	\caption{Architecture of SalNavNet. }
	\label{fig:SalNavNet}
\end{figure*}

Existing image saliency models cannot effectively find the correlation between consecutive frames in a dynamically changing environment. Simultaneously, the existing video saliency models (such as SalEMA, SalLSTM) can learn the video context relationship with the help of structures such as ConvLSTM and  Exponential Moving Average(EMA). Although video saliency models perform better than image saliency models on most saliency metrics, some factors affect the attention generated by the video saliency model to improve the accuracy of Attention-SLAM.
  
 Due to the impact of the saliency data collection method, most saliency datasets have a strong center bias. Most state-of-the-art visual saliency models also learn this center bias from the datasets, making visual saliency models more inclined to focus on the central area of images and ignore the non-central area of the images. In the frame sequences, the positions of salient objects move with the lens. Due to the center bias of existing saliency models, only when these salient objects reach the center of the images, the saliency model marks it as a salient area. When these objects move to the edge of the image, the saliency model ignores these objects. This shift of attention may imitate human eye saccades, making these models perform well on saliency benchmarks. However, the shift of attention makes the visual SLAM system unable to consistently focus on the same salient features. Focusing on the same salient features is the key to improve the accuracy of Attention-SLAM. Therefore, we hope to design a network to avoid rapid changes in attention (eye saccades) while focusing on contextual information. In Attention-SLAM, we hope the saliency model to focus on the same feature points continuously, whether or not they are in the center of the image.
 
Linardos et al. \cite{8} proposed SalEMA and SalLSTM based on SalGAN. These two networks utilize the generator part of SalGAN and use two structures, EMA and ConvLSTM, which enable networks to learn the correlation between frames. Surprisingly, the simple structure of SalEMA is better than SalLSTM. Xinyi Wu et al. \cite{52} propose a correlation-based ConvLSTM, which assigns different weights to consecutive image frames. Inspired from this work, we propose a visual saliency network called SalNavNet that can better improve the accuracy of Attention-SLAM.
\begin{figure}[hbtp]
	\centering	
	\includegraphics[width=0.9\linewidth]{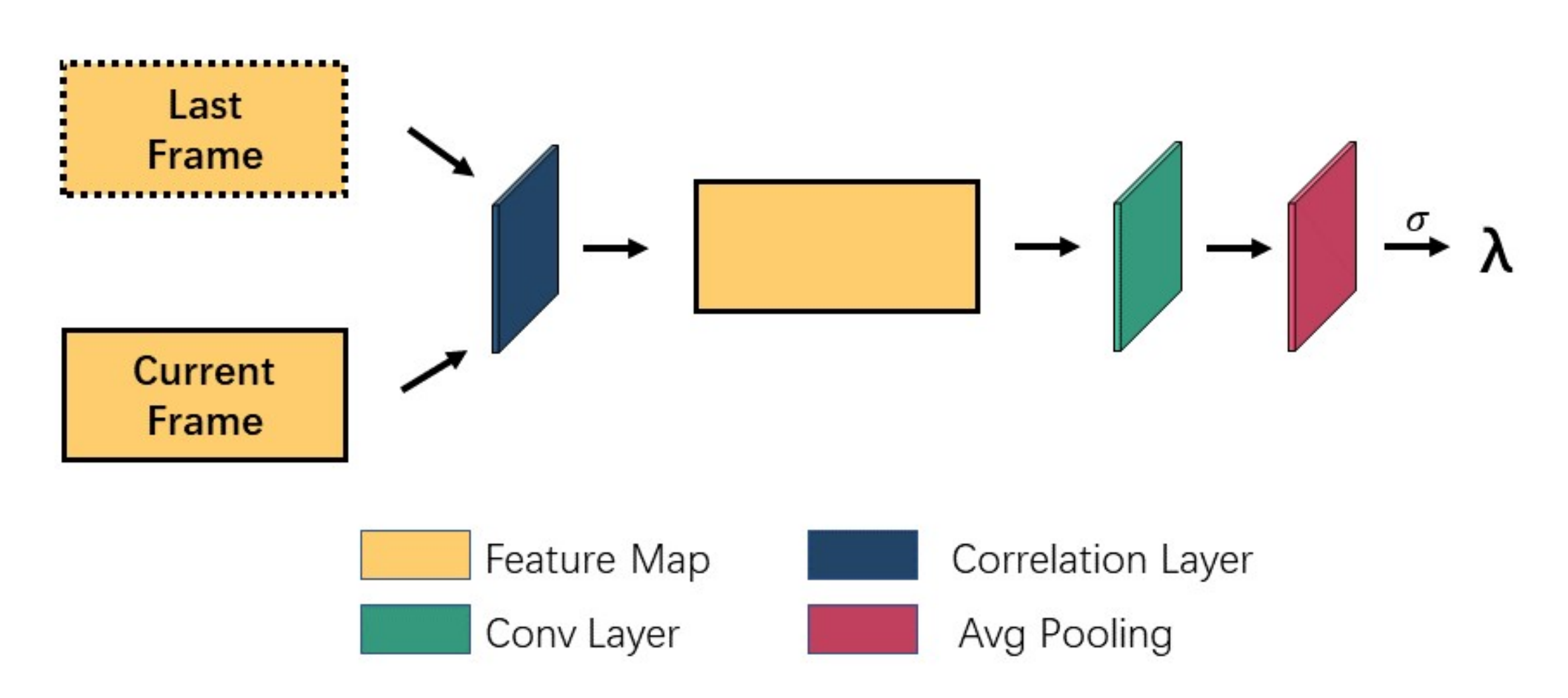}
	
	\caption{Architecture of frame correlation module.}
	\label{fig:Correlation}
\end{figure}
\begin{figure}[hbtp]
	\centering	
	\includegraphics[width=0.7\linewidth]{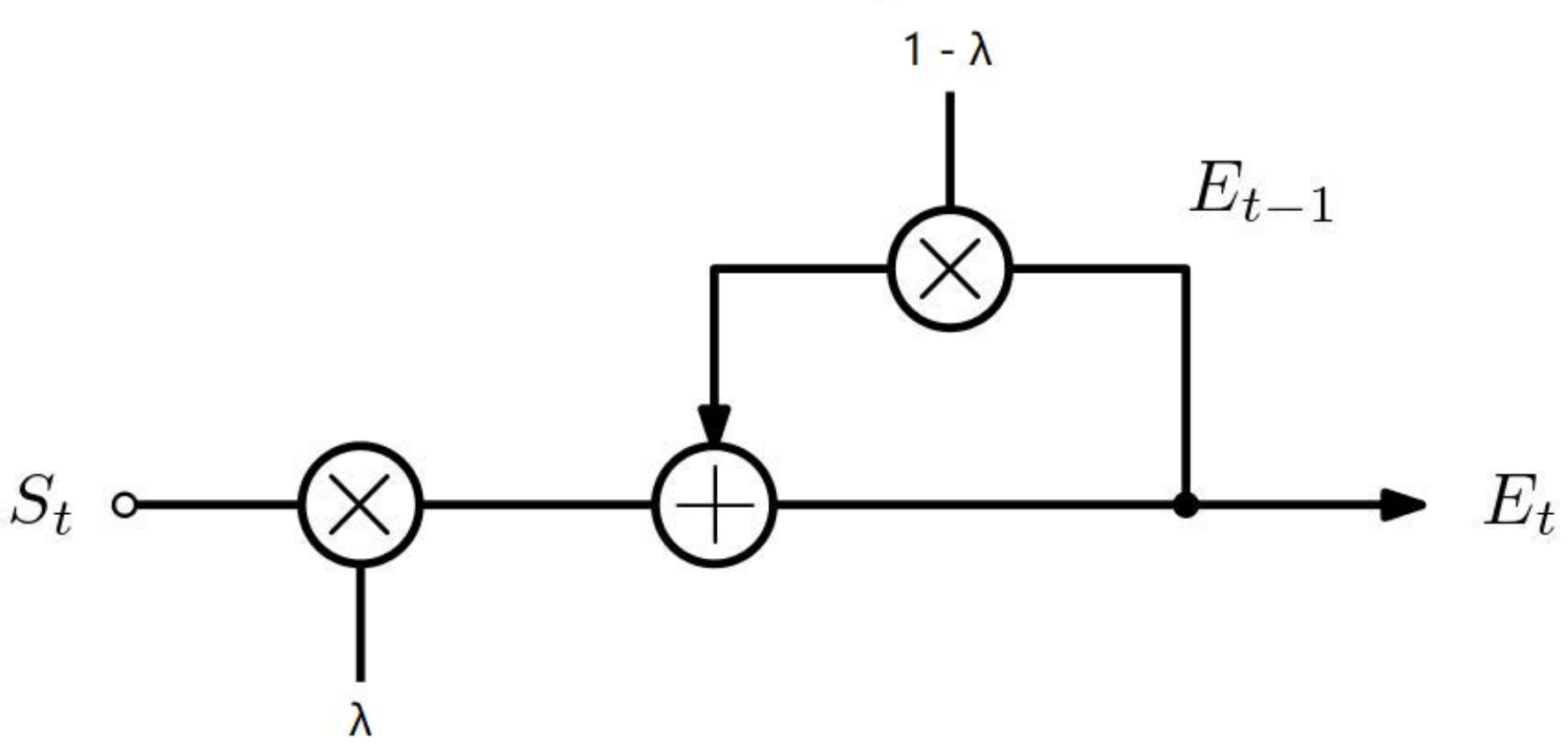}
	
	\caption{Architecture of adaptive exponential moving average module.}
	\label{fig:EMAweighted.png}
\end{figure}

The network structure of SalNavNet is shown in Fig. \ref{fig:SalNavNet}. It applies the same encoder and decoder network structure as SalEMA and SalGAN. The encoder part is a VGG-16 \cite{15} convolutional network. The decoder part applies the same layers as in the encoder in reverse order while using the upsampling operation instead of the pooling operation. To learn the continuous information between frames, we first utilize the frame correlation module to compare the feature map of the current frame through the encoder output and the feature map of the previous frame through the encoder output. The structure of frame correlation module is shown in Fig. \ref{fig:Correlation}. Finally, we get the correlation coefficient $\lambda$ of the two frames, and the correlation coefficient is passed into the adaptive EMA module.

In the frame correlation module, we use the correlation layer proposed in Flownet \cite{53} to compare the similarity of the feature maps of the two adjacent frames and calculate the similarity coefficient $\lambda$. When $\lambda$ is close to 1, it means there is no change in the two feature maps. When the difference of adjacent feature maps is larger, it will make the value of $\lambda$ smaller. Therefore, when there is a huge change between the two adjacent feature maps, the saliency map generated by the saliency model has a rapid change of attention. We designed an adaptive EMA module, as shown in Fig. \ref{fig:EMAweighted.png}. On one hand, the adaptive EMA module allows our model to learn continuous information between frames. On the other hand, the introduction of the similarity coefficient $\lambda$ mitigates the center bias of saliency models and rapid change of attention. In the field of visual saliency, the rapid changes of attention can better imitate the real data of the dataset. However, in SLAM, we hope that the model continue to focus on the same salient target in the environment.

The parameters of the encoder-decoder convolutional layers are adopted from SalEMA \cite{8}. We train our model on DHF1K dataset with 5 epochs. The DHF1K dataset contains 700 videos. BCE function is adopted as the training loss function, the formula is as follows:
\begin{equation}
	L_{B C E}=-\frac{1}{N} \sum_{n=1}^{N} P_{n} \log \left(Q_{n}\right)+\left(1-P_{n}\right) \log \left(1-Q_{n}\right)
\end{equation}
where $P_{n}$ represents the $n$-th generated saliency map and $Q_{n}$ represents the $n$-th ground-truth saliency map. We use the Adam optimizer with a learning rate of $10^{-7}$.

\subsection{Weighted Bundle Adjustment}
Bundle Adjustment (BA) is a method for solving the least-squares problem, which gives the same weight to each feature point in the estimation process. After that, we use the model generated by the visual saliency model as the weight value. The saliency maps are grayscale maps, where white parts have a value of 255 and black parts have a value of 0. In order to use the saliency maps as weights, we normalize the maps as weights shown in the equation (\ref{equation:Weighted Bundle adjustment}).

\begin{equation}
\begin{array}{c}
\left\{\boldsymbol{X}^{i}, \boldsymbol{R}_{\iota}, \boldsymbol{t}_{l} | i \in \boldsymbol{P}_{\mathrm{L}}, l \in \boldsymbol{T}_{\mathrm{L}}\right\}= \\
\mathop{\arg\min}_{\boldsymbol{X}^{i}, \boldsymbol{R}_{l}, \boldsymbol{t}_{l}} \sum_{j \in T_{L} \cup T_{F}}\sum_{k \in \chi_{j}}\boldsymbol{\omega} \rho(E(j, k))
\end{array}\\
\label{equation:Weighted Bundle adjustment}
\end{equation}
where $\boldsymbol{P}_{\mathrm{L}}$ means the pose of the current keyframe and local keyframes which has a co-view relationship with the current keyframe. $\boldsymbol{T}_{\mathrm{L}}$ means the three-dimensional coordinates of all map points that can be observed in the local keyframes. $\rho$ means the Huber function \cite{54}. $\chi_{j}$ is the set of feature points matched to the map point for the $j$-th local keyframe. $E(j, k)$ is defined as:
 \begin{equation}
 	E(j, k)=\left\|\boldsymbol{x}_{\mathrm{m}}^{k}-\pi_{\mathrm{m}}\left(\boldsymbol{R}_{j} \boldsymbol{X}^{k}+\boldsymbol{t}_{j}\right)\right\|_{\boldsymbol{\Sigma}}^{2}
 	\label{equation:BA error}
 \end{equation}
However, sometimes the saliency area in the saliency map is small. If we directly normalize these maps, it may cause tracking failure. So we add a constant factor $b$ when we normalize the weight. The normalizing operation is defined as:
\begin{equation}
	\omega_{i} = \dfrac{p_{i} + b}{255}
	\label{equation: weight normalization}
\end{equation}
where $\omega_{i}$ is the weight corresponding with the $i$-th feature point. $ p_{i} $ is the pixel value in the saliency maps corresponding to every feature point.
\subsection{Entropy Based KeyFrame Selection and Evaluation}
In this sub-section, we first evaluated the uncertainty of motion estimation from the perspective of information theory, and then we used the concept of entropy reduction as the criterion for selecting keyframes to further improve the performances of Attention-SLAM.
\subsubsection{KeyFrame Selection Using Entropy Ratio}
During the motion estimation in Attention-SLAM, equation (\ref{equation:BA error}) is used to calculate the reprojection error. Furthermore, we want to know the uncertainty between relative motion estimation. C. Kerl et al. \cite{36} prove that the uncertainty of relative motion estimation can be measured by the covariance matrix described in equation (\ref{equation: covirance}).
\begin{equation}
\label{equation: covirance}
\boldsymbol{\xi}_{t, t+1} \sim \mathcal{N}\left(\boldsymbol{\xi}_{t, t+1}^{*}, \boldsymbol{\Sigma}_{\xi_{t, t+1}}^{*}\right)
\end{equation}
where $\boldsymbol{\xi}_{t, t+1}$ is the vector of the camera motion between keyframes $t$ and $t + 1$, $\boldsymbol{\Sigma}_{\xi_{t, t+1}}^{*}$ is the covariance of the estimated motion approximated by the inverse of the Hessian of the cost function in the last iteration \cite{35}.  We use $\boldsymbol{\Sigma}_{\xi_{t, t+1}}^{*}$ to calculate the uncertainty of motion estimation. The entropy of a keyframe which has $m$ dimensions can be defined as:  
\begin{equation}
	H(\boldsymbol{x})=0.5 m(1+\ln (2 \pi))+0.5 \ln (|\boldsymbol{\Sigma}|)
	\label{equation: differential entropy}
\end{equation}
Equation (\ref{equation: differential entropy}) shows that the determinant of the covariance matrix is proportional to the entropy, ($H(\boldsymbol{x}) \propto \ln (|\mathbf{\Sigma}|)$). Equation (\ref{equation: differential entropy}) is used to represent uncertainty. In ORB-SLAM \cite{7}, several metrics are used to select keyframes. However, when scenes changes, simple keyframe selection metrics may not perform well. We compute the entropy ratio $\alpha$ \cite{36}\cite{37} between the motion  estimate $\xi_{k, k+j}$ from the last keyframe $k$ to the current frame $j$ and the motion estimation  $\xi_{k, k+j}$ from keyframe $k$ to the next frame $k + 1$:
\begin{equation}
	\alpha=\frac{H\left(\boldsymbol{\xi}_{k: k+j}\right)}{H\left(\boldsymbol{\xi}_{k: k+1}\right)}
	\label{equation: entropy ratio}
\end{equation}  
We set the threshold of $\alpha$ as 0.9. When the entropy ratio of a frame is over 0.9, it will not be chosen as a keyframe. Because it means the current frame won't reduce the uncertainty of the motion estimation efficiently. The entropy ratio is similar to the likelihood ratio test for loop closure validation described by Stuckler et al. \cite{38}.
\subsubsection{Entropy Reduction Evaluation}
Our method combines the visual saliency model with the SLAM method. The saliency model extracts semantic saliency information from the environment, this might make the trajectory estimated by our system closer to the ground-truth. Therefore, we analyze the influence of our method on the uncertainty of pose estimation from the perspective of information theory.

Firstly we calculate the sum of all the determinant of keyframes covariance matrix in each keyframe, and then we use it to find the average uncertainty of each keyframe $\beta$ in the SLAM method. The average entropy is shown in equation (\ref{equation: average entropy}).
\begin{equation}
\beta = \frac{\sum_{i=1}^{n - 1}{\boldsymbol{\Sigma}_{\xi_{t, t+1}}^{*}}}{n} 
\label{equation: average entropy}
\end{equation}

where $n$ denotes the number of keyframes. We compute the entropy reduction $\gamma$ \cite{31}\cite{32} between ORB-SLAM and Attention-SLAM. If our method has less uncertainty than ORB-SLAM during the pose estimation process, the $\gamma$ will larger than zero. 
\begin{equation}
\gamma = \log_2 \frac{\beta_{ORB SLAM}}{\beta_{Attention-SLAM}}
\label{equation: entropy reduction}
\end{equation}
\section{Experiment}
We first analyze the impact of the saliency maps generated by different saliency models on the Attention-SLAM and use saliency models to generate a novel saliency dataset called Salient Euroc. Then we compare our method with other state-of-the-art visual SLAM methods on the Salient Euroc dataset. Furthermore, the entropy reduction of Attention-SLAM is analyzed and the entropy reduction is utilized as a criterion for selecting keyframes. Our approach is compared with the baseline method on a device with i5-9300H CPU(2.4 GHz) and 16G RAM. 
\subsection{Image Saliency Models on Attention-SLAM}
\begin{figure}[htbp]
	\centering	
	\subfloat[]{
		\begin{minipage}[t]{0.3\linewidth}
			\centering
			\includegraphics[clip,width=\columnwidth]{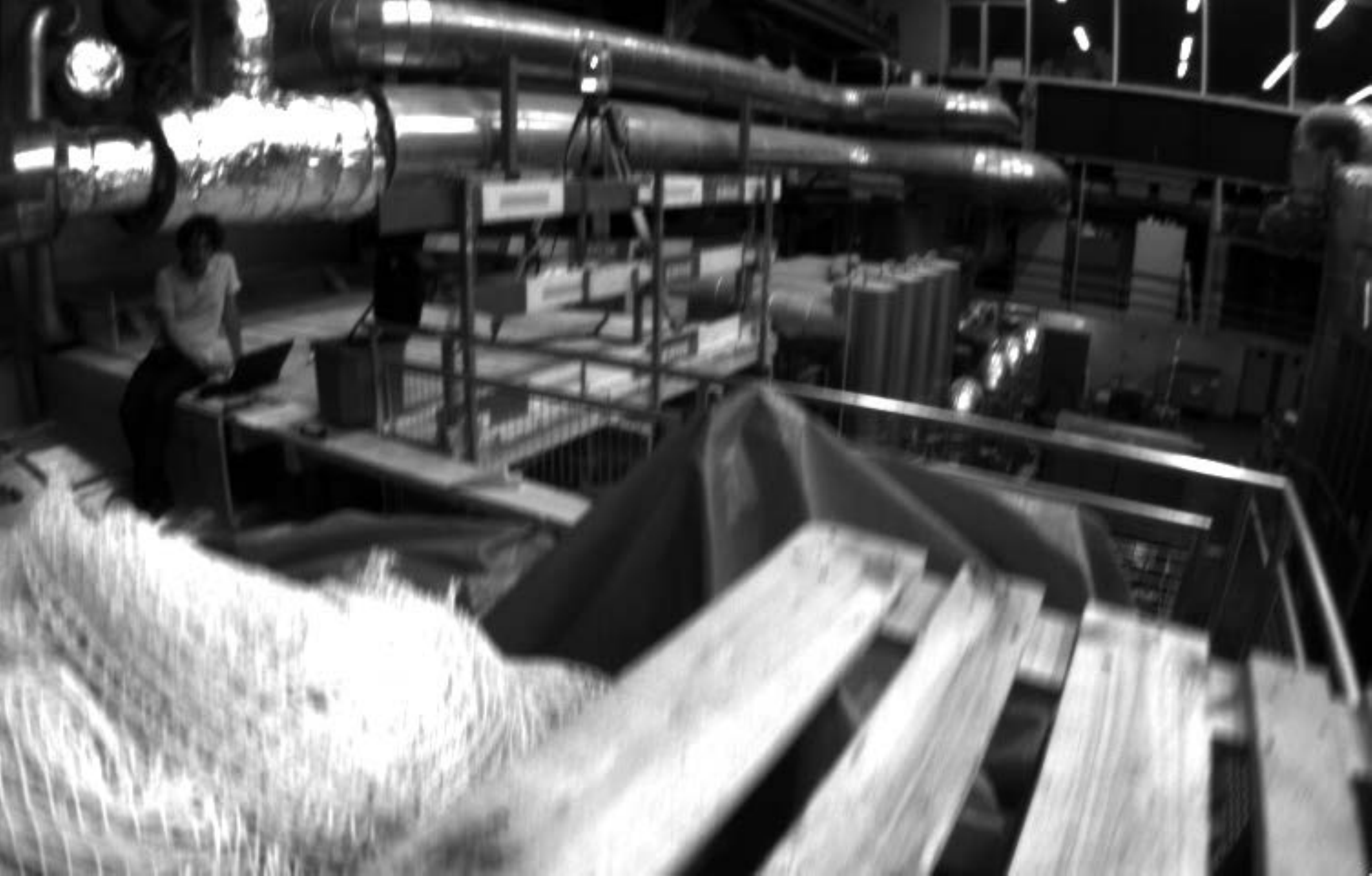}
			\centering
			\includegraphics[clip,width=\columnwidth]{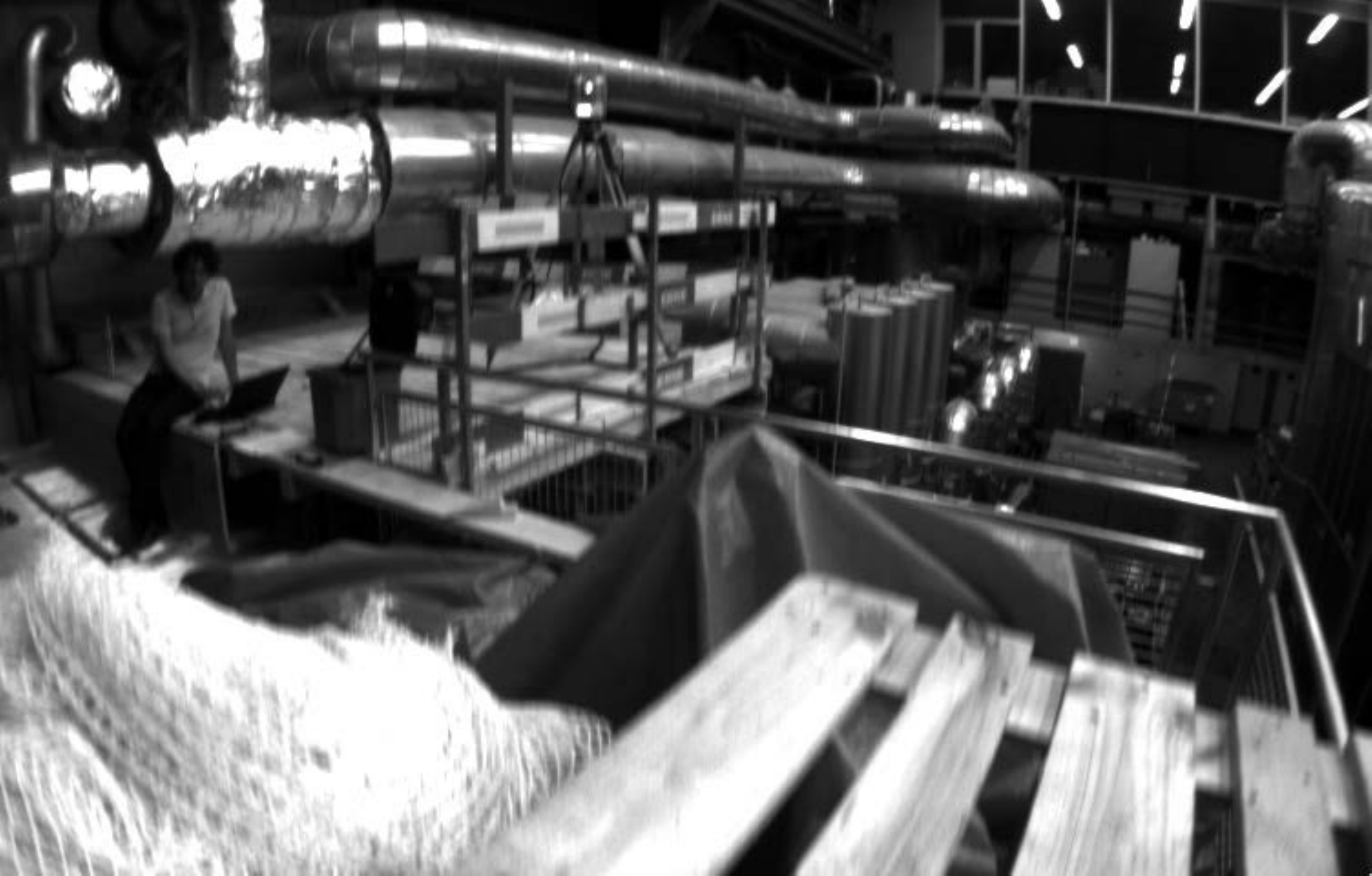}
			\centering
			\includegraphics[clip,width=\columnwidth]{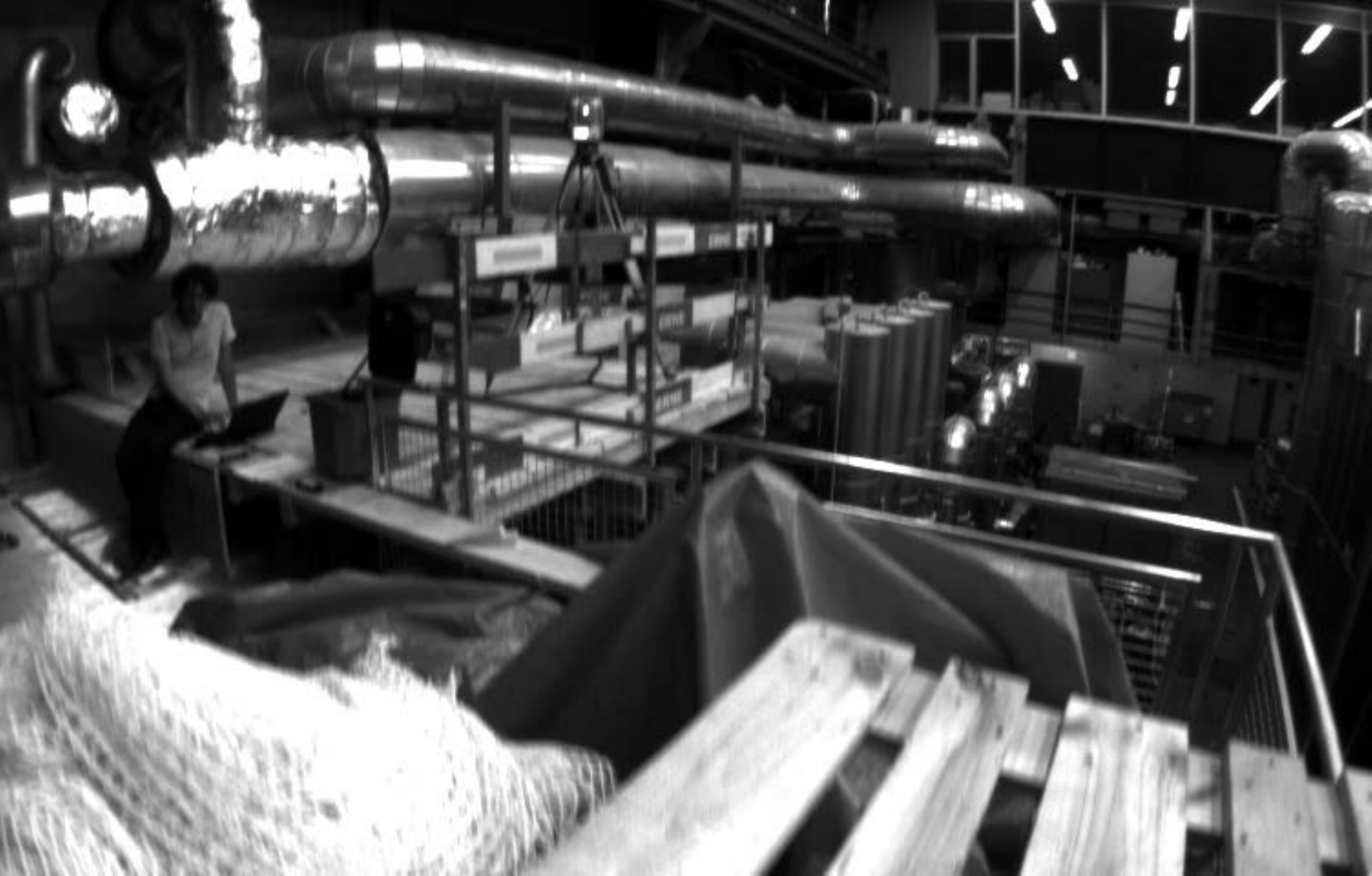}
		\end{minipage}%
	}%
	\subfloat[]{
		\begin{minipage}[t]{0.3\linewidth}
			\centering
			\includegraphics[clip,width=\columnwidth,height=0.645\columnwidth]{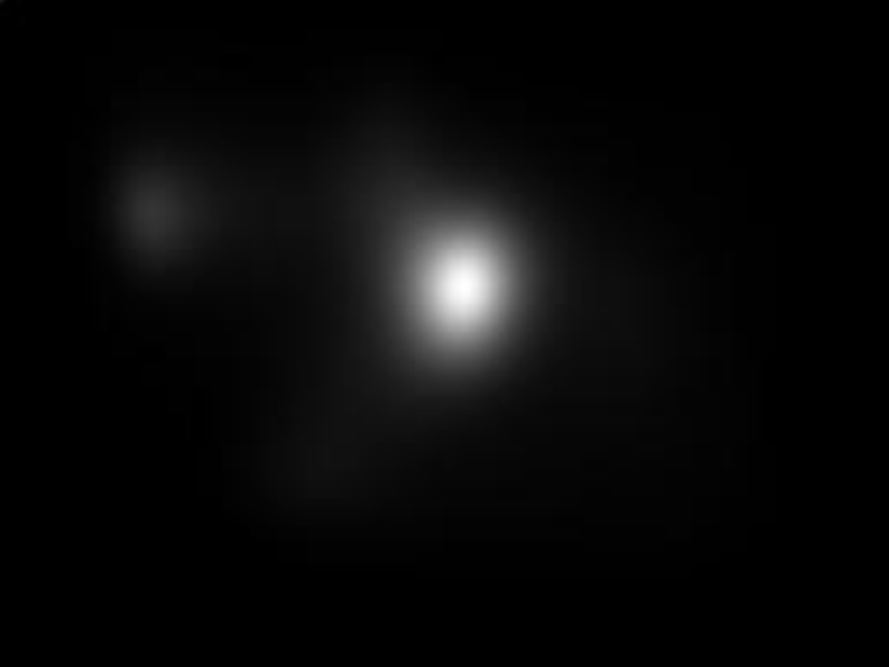}
			\centering
			\includegraphics[clip,width=\columnwidth,height=0.645\columnwidth]{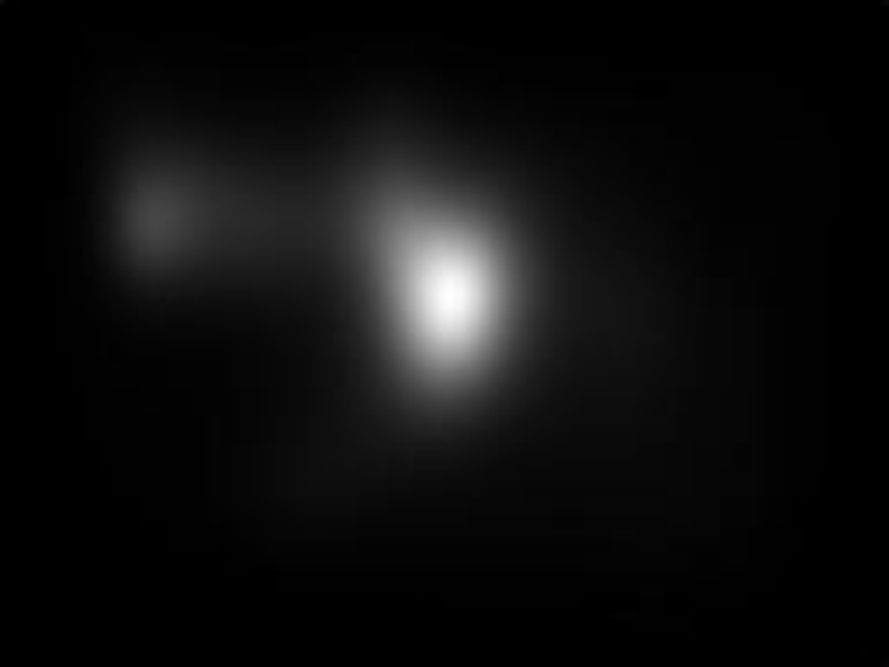}
			\centering
			\includegraphics[clip,width=\columnwidth,height=0.645\columnwidth]{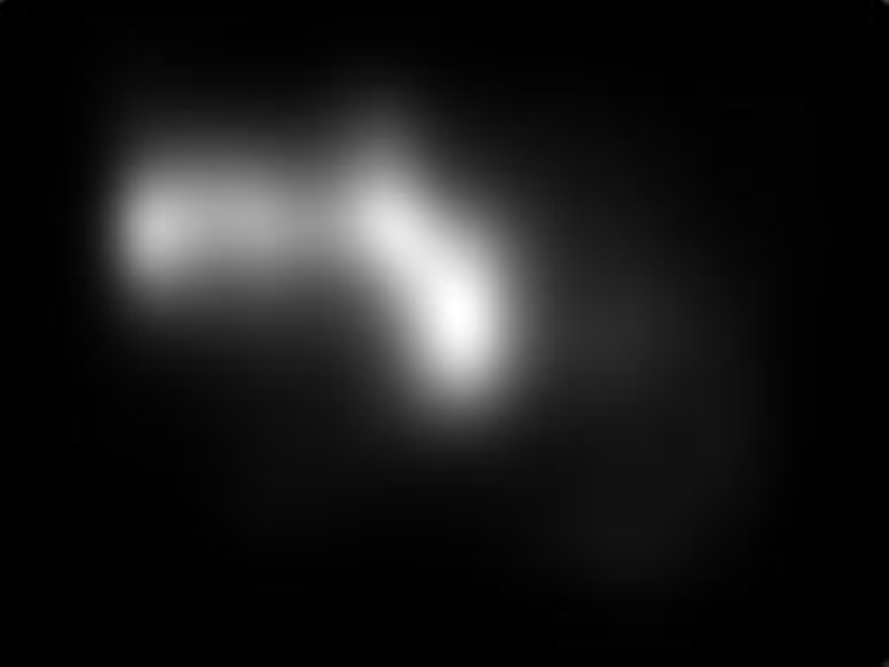}
		\end{minipage}%
	}%
	\subfloat[]{
		\begin{minipage}[t]{0.3\linewidth}
			\centering
			\includegraphics[clip,width=\columnwidth]{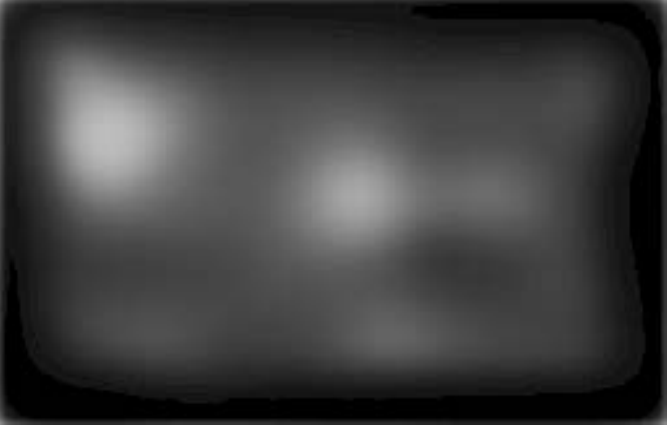}
			\centering
			\includegraphics[clip,width=\columnwidth]{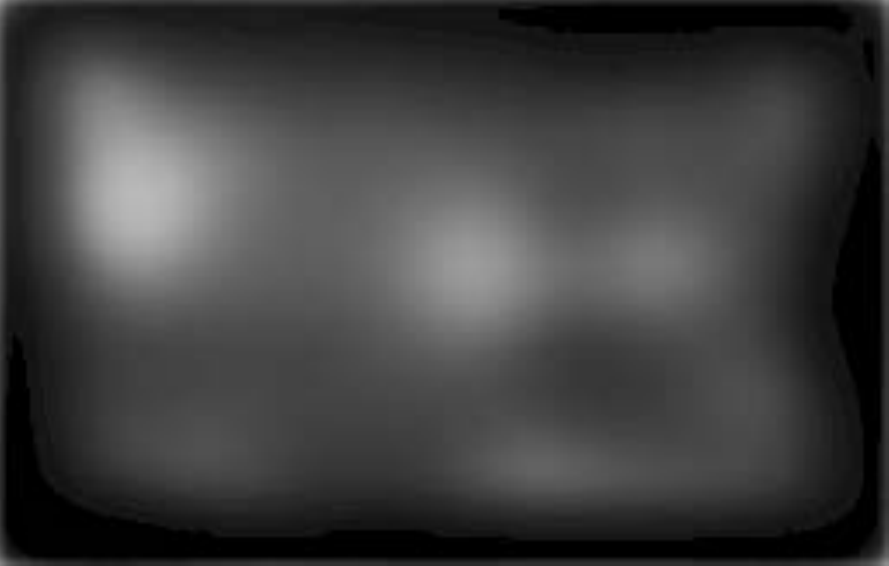}
			\centering
			\includegraphics[clip,width=\columnwidth]{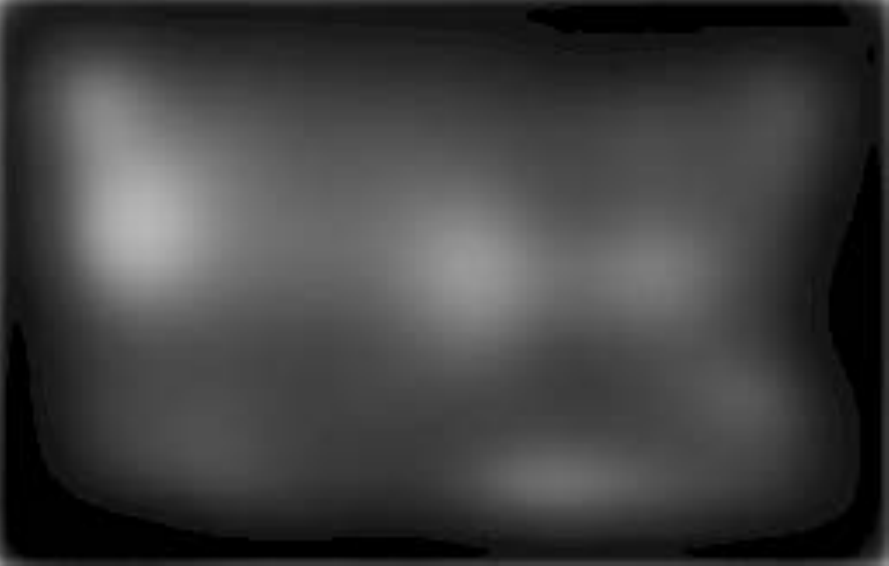}
		\end{minipage}%
		
	}%
	
	\centering
	\caption{Saliency map comparison: (a) Original image sequences (b) Saliency map gnerated by SalEMA (c) Saliency map generated by SalGAN}
	\label{fig: Comparison of SalGAN and SalEMA}
\end{figure}

We select the image saliency model SalGAN to compare the state-of-the-art saliency model SalEMA. The saliency maps corresponding to the Euroc dataset is generated. As shown in Fig. \ref{fig: Comparison of SalGAN and SalEMA}, the saliency maps generated by the two models are different. The saliency area in the saliency maps generated by SalEMA is small. The center bias of the saliency maps generated by SalGAN is weak. In Tab. \ref{table: Comparison between different saliency models}, we calculate the Root Mean Square (RMSE) of Absolute Trajectory Error (ATE). The saliency maps generated by SalGAN help Attention-SLAM perform better in most data sequences. This result verifies our previous conjecture. Saliency maps with weak center bias enable Attention-SLAM to achieve higher accuracy.
We use the saliency maps generated by SalGAN and SalEMA as weights into Attention-SLAM and compare the Root Mean Square Error (RMSE) of Absolute Trajectory Error (ATE). The result is shown in Tab. \ref{table: Comparison between different saliency models}
\begin{table}[]
	\centering
	\caption{RMSE of Absolute Trajectory Error between ORB-SLAM and Attention-SLAM using weights generated by different saliency models. (Unit: meter)}
	\label{table: Comparison between different saliency models}
	\begin{tabular}{ccc}
		\hline
		Seq.Name&  Our Method(SalGAN) & Our Method(SalEMA) \\ \hline
		MH01 & \textbf{0.042352}      & 0.0437                  \\
		MH02 & \textbf{0.034408}      & 0.03776                 \\
		MH03 & \textbf{0.035663}      & 0.05871                 \\
		MH04 & 0.094429               & \textbf{0.0763}         \\
		MH05 & \textbf{0.046727}      & 0.04895                 \\
		V101 & \textbf{0.095468}      & 0.09615                 \\
		V102 & \textbf{0.063389}      & 0.06496                 \\
		V103 & \textbf{0.068524}      & 0.0767                  \\
		V201 & 0.057869               & \textbf{0.0578}         \\
		V202 & 0.058027               & \textbf{0.0566}         \\
		V203 & \textbf{0.105576}      & 0.11933                 \\ \hline
	\end{tabular}
\end{table}
The results indicate that the saliency maps generated by SalGAN help our method achieve better performance in most data sequences.

\subsection{Comparison of Vedio Saliency Model and SalNavNet}
In the video saliency model, Linardos et al. \cite{8} introduced ConvLSTM and EMA two structures to learn the correlation between adjacent frames. SalEMA achieved state-of-the-art performance on the visual saliency evaluation metrics. Besides, to help the SLAM system achieve a better performance, we consider the related information of the consecutive frames and make the saliency model pay attention to the same object continuously. 
\begin{figure}[htbp]
	\centering	
	\subfloat[]{
		\begin{minipage}[t]{0.3\linewidth}
			\centering
			\includegraphics[clip,width=\columnwidth]{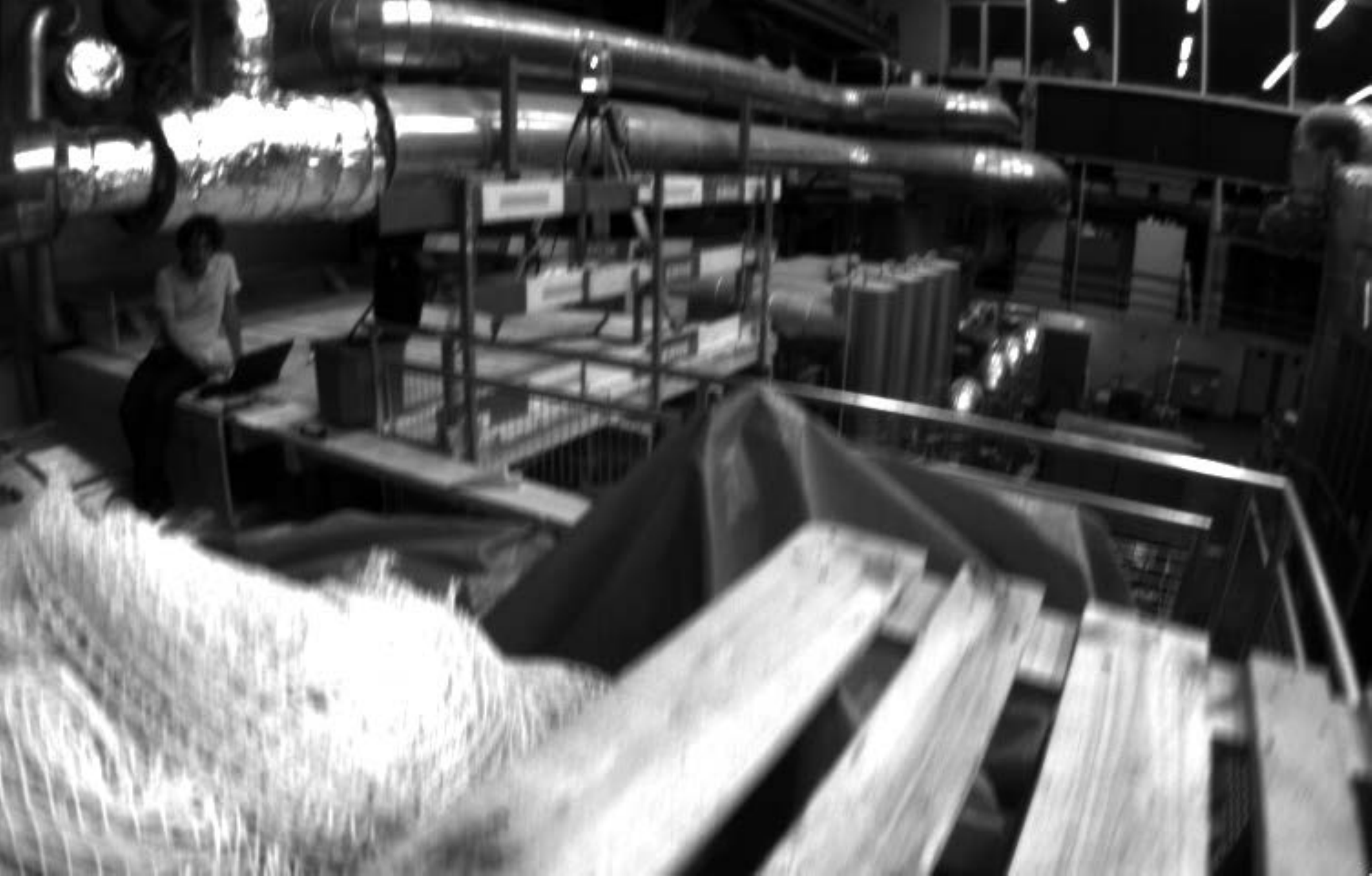}
			\centering
			\includegraphics[clip,width=\columnwidth]{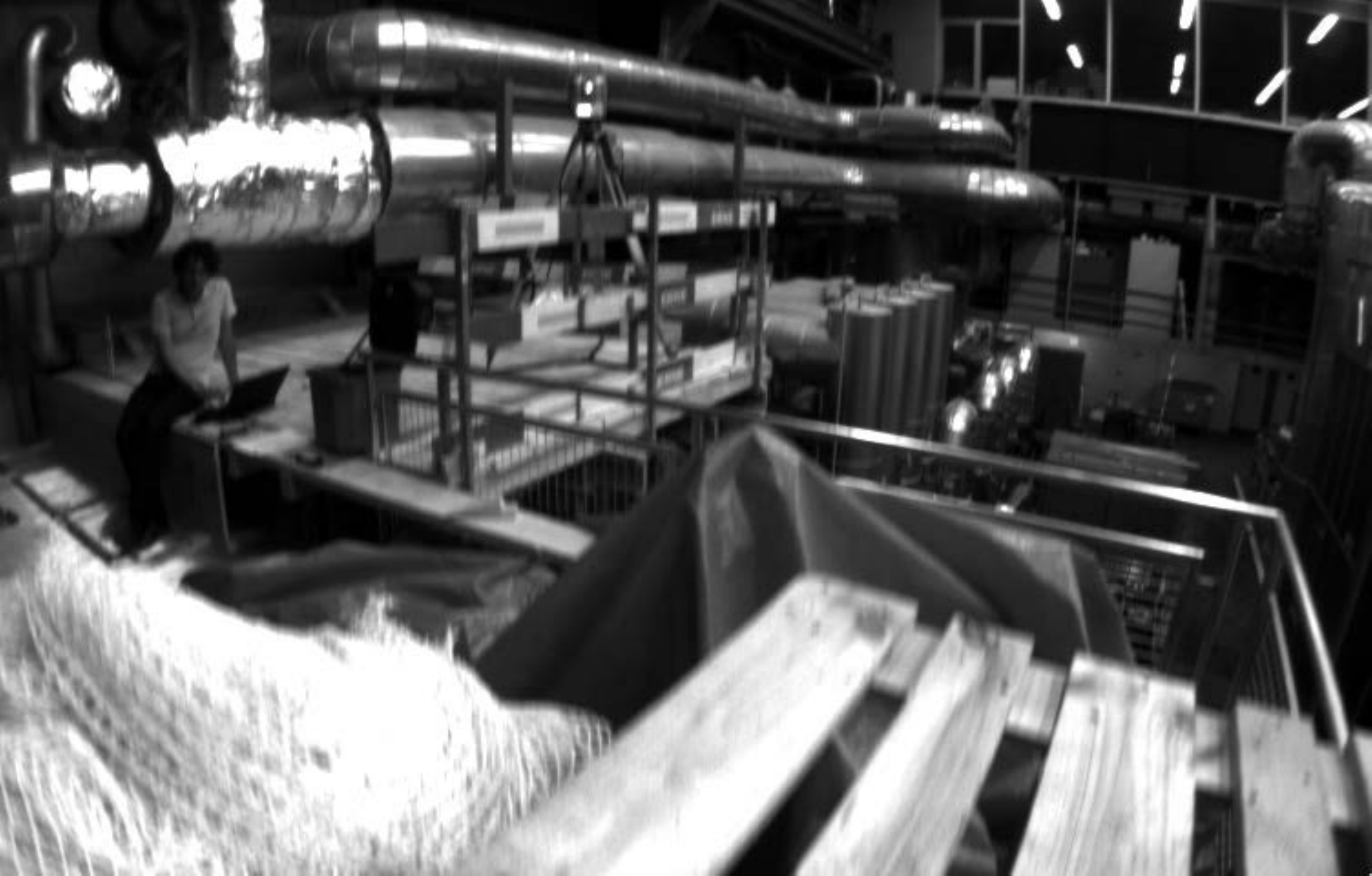}
			\centering
			\includegraphics[clip,width=\columnwidth]{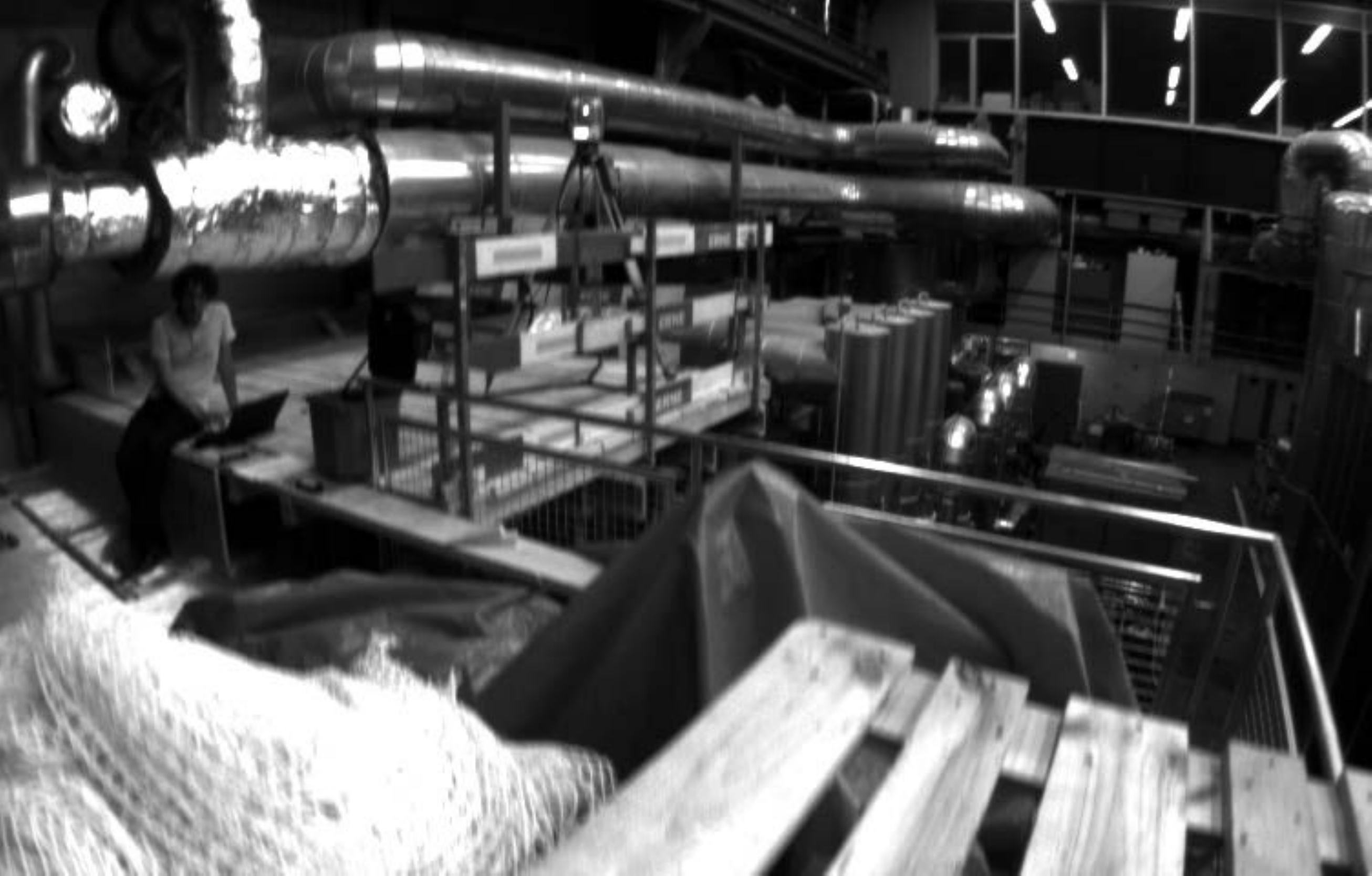}
		\end{minipage}%
	}%
	\subfloat[]{
		\begin{minipage}[t]{0.3\linewidth}
			\centering
			\includegraphics[clip,width=\columnwidth,height=0.645\columnwidth]{li10.pdf}
			\centering
			\includegraphics[clip,width=\columnwidth,height=0.645\columnwidth]{li11.pdf}
			\centering
			\includegraphics[clip,width=\columnwidth,height=0.645\columnwidth]{li12.pdf}
		\end{minipage}%
	}%
	\subfloat[]{
		\begin{minipage}[t]{0.3\linewidth}
			\centering
			\includegraphics[clip,width=\columnwidth,height=0.645\columnwidth]{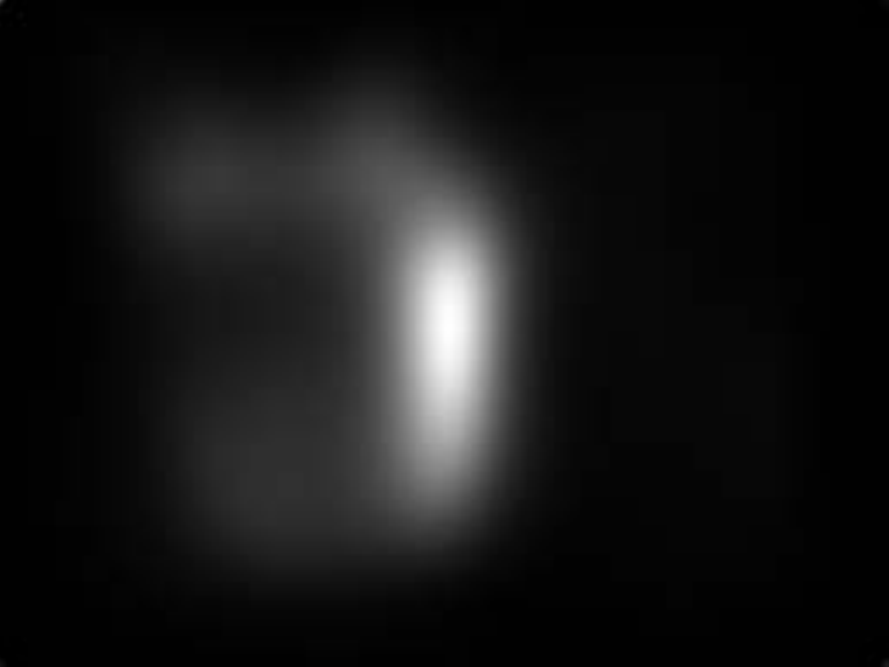}
			\centering
			\includegraphics[clip,width=\columnwidth,height=0.645\columnwidth]{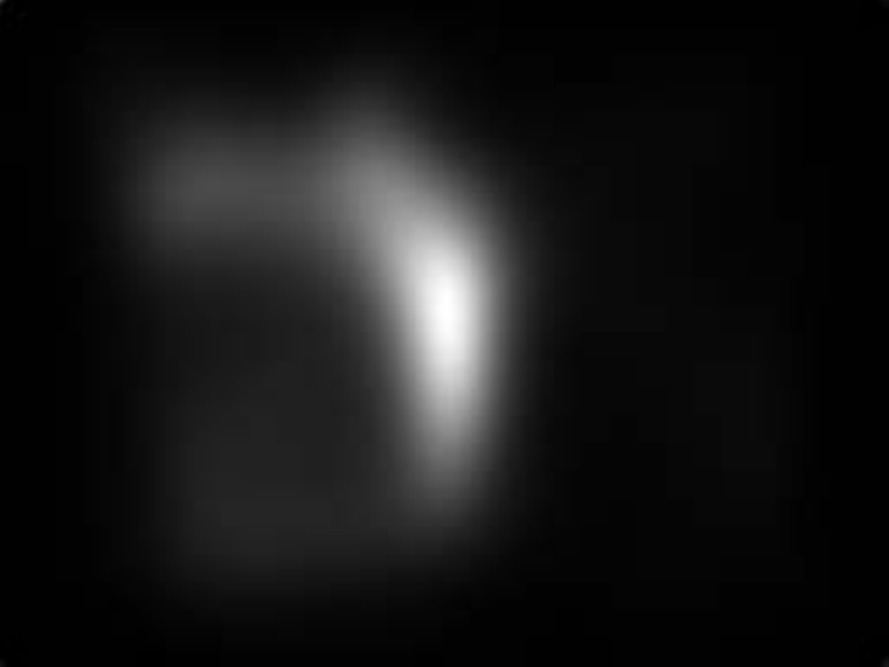}
			\centering
			\includegraphics[clip,width=\columnwidth,height=0.645\columnwidth]{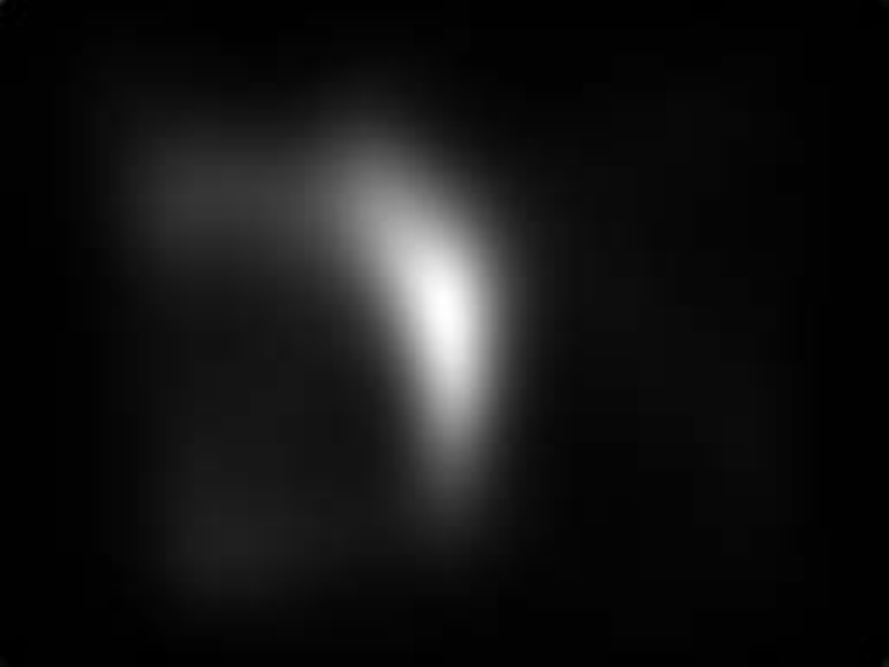}
		\end{minipage}%
		
	}%
	
	\centering
	\caption{Saliency map comparison: (a) Original image sequences (b) Saliency map gnerated by SalEMA (c) Saliency map generated by SalNavNet.}
	\label{fig: Comparison of SalEMA and SalNav}
\end{figure}
Fig. \ref{fig: Comparison of SalEMA and SalNav} compares the saliency maps generated by the two models. The saliency maps generated by the SalEMA has a strong center bias. Although the three adjacent original images has little change, the saliency map generated by SalEMA has significantly changed. From Fig. \ref{fig: Comparison of SalEMA and SalNav} (c), we can find that the saliency maps generated by SalNavNet mitigate the center bias. The results in Tab. \ref{table: Comparison between SalNav and SalEMA} shows that SalNavNet performs better than SalEMA in the most data sequences. It means that the saliency maps generated by SalNavNet can help Attention-SLAM perform better than SalEMA. 
\begin{table}[]
	\centering
	\caption{RMSE of Absolute Trajectory Error between Attention-SLAM using state of the art saliency model or using SalNavNet. (Unit: meter) }
	\label{table: Comparison between SalNav and SalEMA}
	\begin{tabular}{ccc}
		\hline
		Seq.Name  &  Our Method(SalEMA)  & Our Method(SalNavNet) \\ \hline
		MH01  & 0.043697             & 0.044992                  \\
		MH02  & 0.037755             & \textbf{0.034467}                 \\
		MH03  & 0.058708             & \textbf{0.037102}                 \\
		MH04  & 0.076253             & \textbf{0.057290}         \\
		MH05  & 0.048950             & \textbf{0.047243}                 \\
		V101  & 0.096147             & \textbf{0.095110}                 \\
		V102  & 0.064959             & \textbf{0.063080}                 \\
		V103  & 0.076695             & 0.101995                  \\
		V201  & 0.057833             & \textbf{0.057760}         \\
		V202  & 0.056589             & \textbf{0.056460}         \\
		V203  & 0.119328             & 0.116147                 \\ \hline
	\end{tabular}
\end{table}
\subsection{Salient Euroc Datasets}
To verify the effectivity of our method, we build a new semantic SLAM dataset called Salient Euroc based on Euroc \cite{9}. Euroc includes eleven datasets and corresponding ground-truth trajectory. These datasets have three levels of difficulty according to texture quality, scene brightness, and speed of motion. Tab. \ref{table:Euroc} shows the difficulty of the data sequences. 
\begin{table}[H]
	\caption{Euroc dataset characteristics \cite{9}}
	\label{table:Euroc}
	\centering
	
	\begin{tabular}{c c c}
		\hline
		Name & Difficulty & Discription                \\ 
		\hline
		MH01 & Easy       & Good texture, bright scene \\
		MH02 & Easy       & Good texture, bright scene \\
		MH03 & Medium     & Fast motion, bright scene  \\
		MH04 & Difficult  & Fast motion, dark scene    \\
		MH05 & Difficult  & Fast motion, dark scene    \\
		V101 & Easy       & Slow motion, bright scene  \\
		V102 & Medium     & Fast motion, bright scene  \\
		V103 & Difficult  & Fast motion, motion blur   \\
		V201 & Easy       & Slow motion, bright scene  \\
		V202 & Medium     & Fast motion, bright scene  \\
		V203 & Difficult  & Fast motion,motion blur    \\ 
		\hline
	\end{tabular}
\end{table}

\begin{figure}[htbp]
	\centering
	
	\subfloat[]{
		\begin{minipage}[t]{0.3\linewidth}
			\centering
			\includegraphics[clip,width=\columnwidth]{li16.pdf}
			\centering
			\includegraphics[clip,width=\columnwidth]{li17.pdf}
			\centering
			\includegraphics[clip,width=\columnwidth]{li18.pdf}
		\end{minipage}%
	}%
	\subfloat[]{
		\begin{minipage}[t]{0.3\linewidth}
			\centering
			\includegraphics[clip,width=\columnwidth, height=0.64\columnwidth]{li19.pdf}
			\centering
			\includegraphics[clip,width=\columnwidth, height=0.64\columnwidth]{li20.pdf}
			\centering
			\includegraphics[clip,width=\columnwidth, height=0.64\columnwidth]{li21.pdf}
		\end{minipage}%
	}%
	\subfloat[]{
		\begin{minipage}[t]{0.3\linewidth}
			\centering
			\includegraphics[clip,width=\columnwidth]{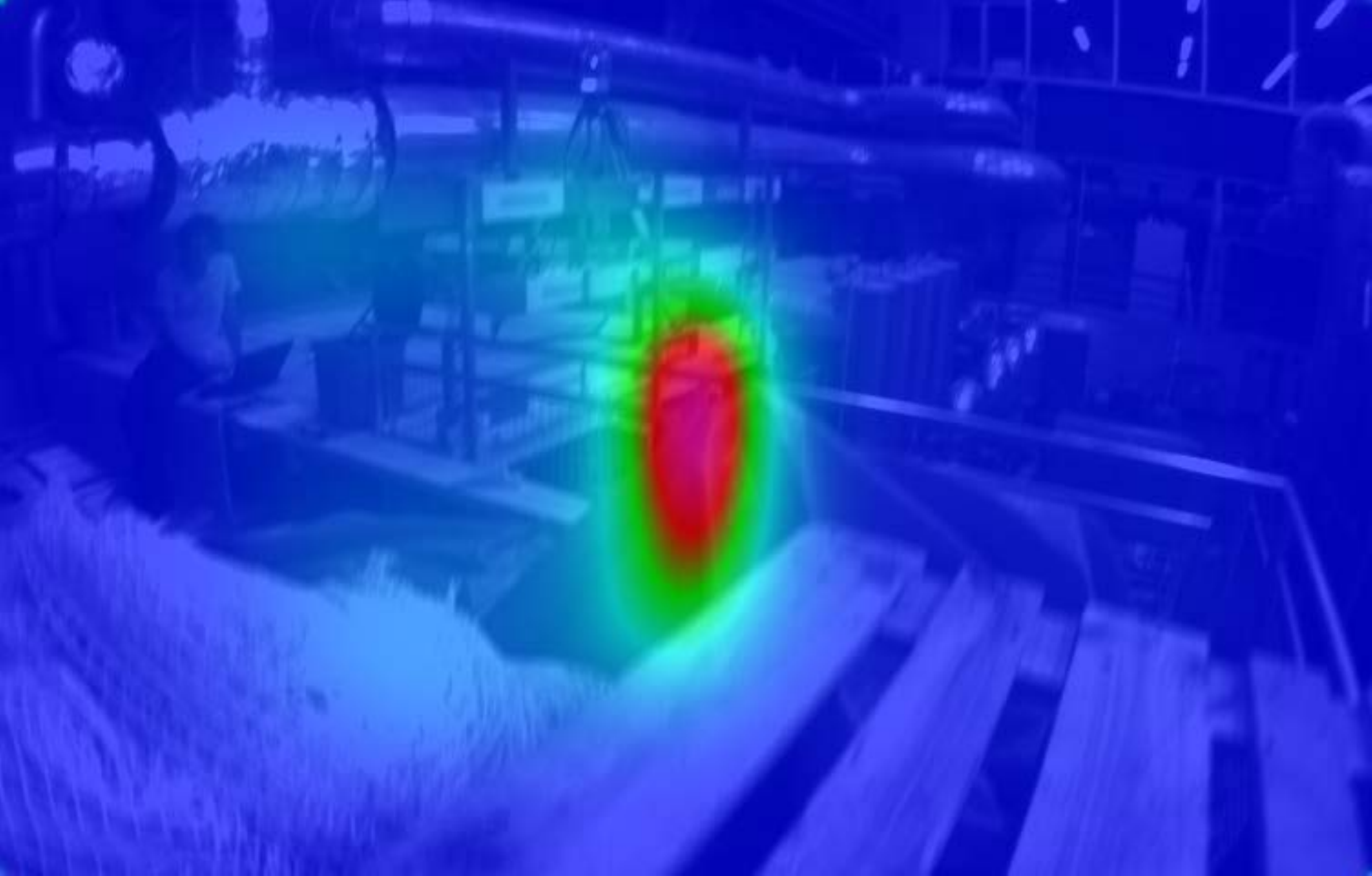}
			\centering
			\includegraphics[clip,width=\columnwidth]{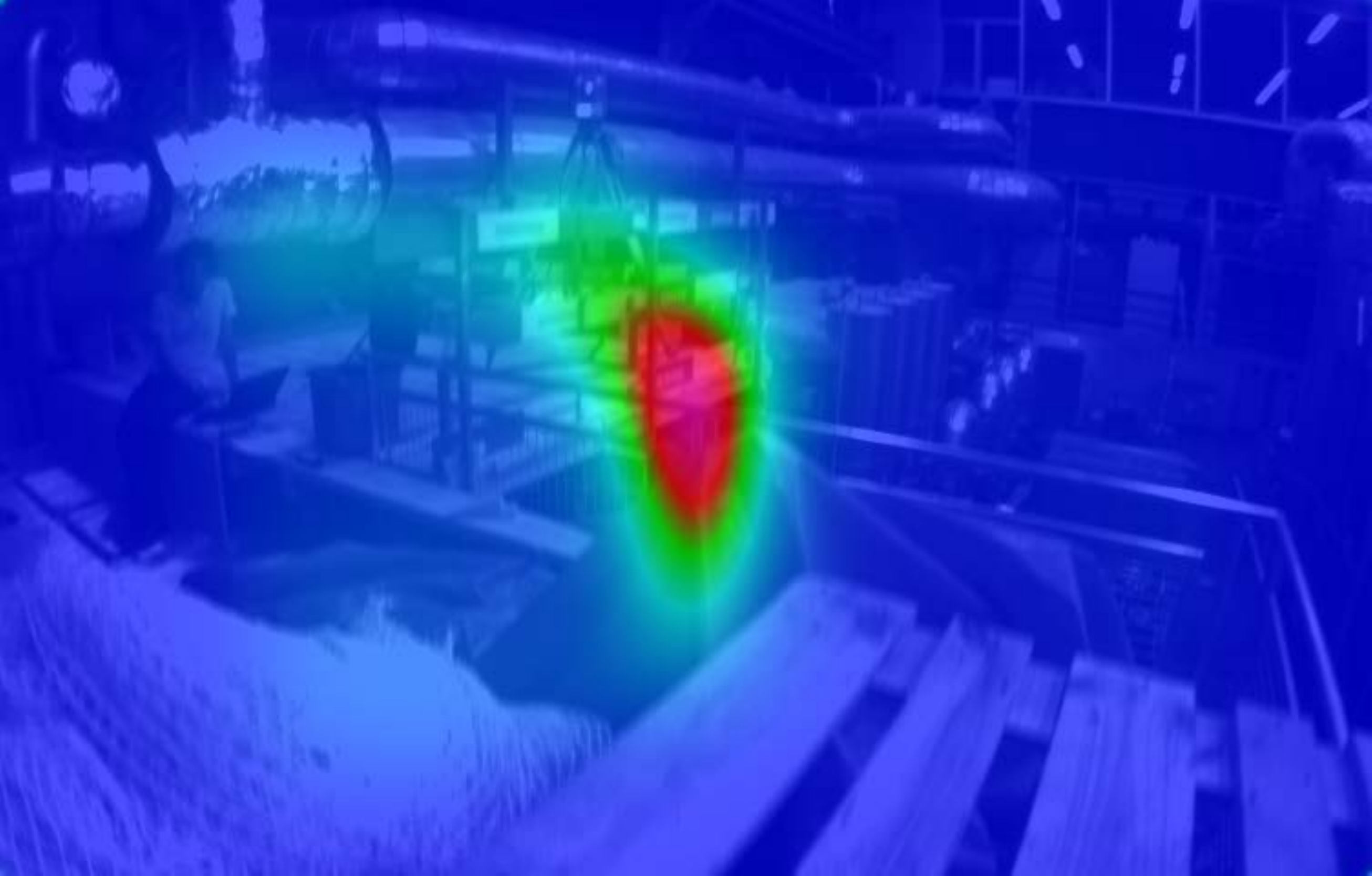}
			\centering
			\includegraphics[clip,width=\columnwidth]{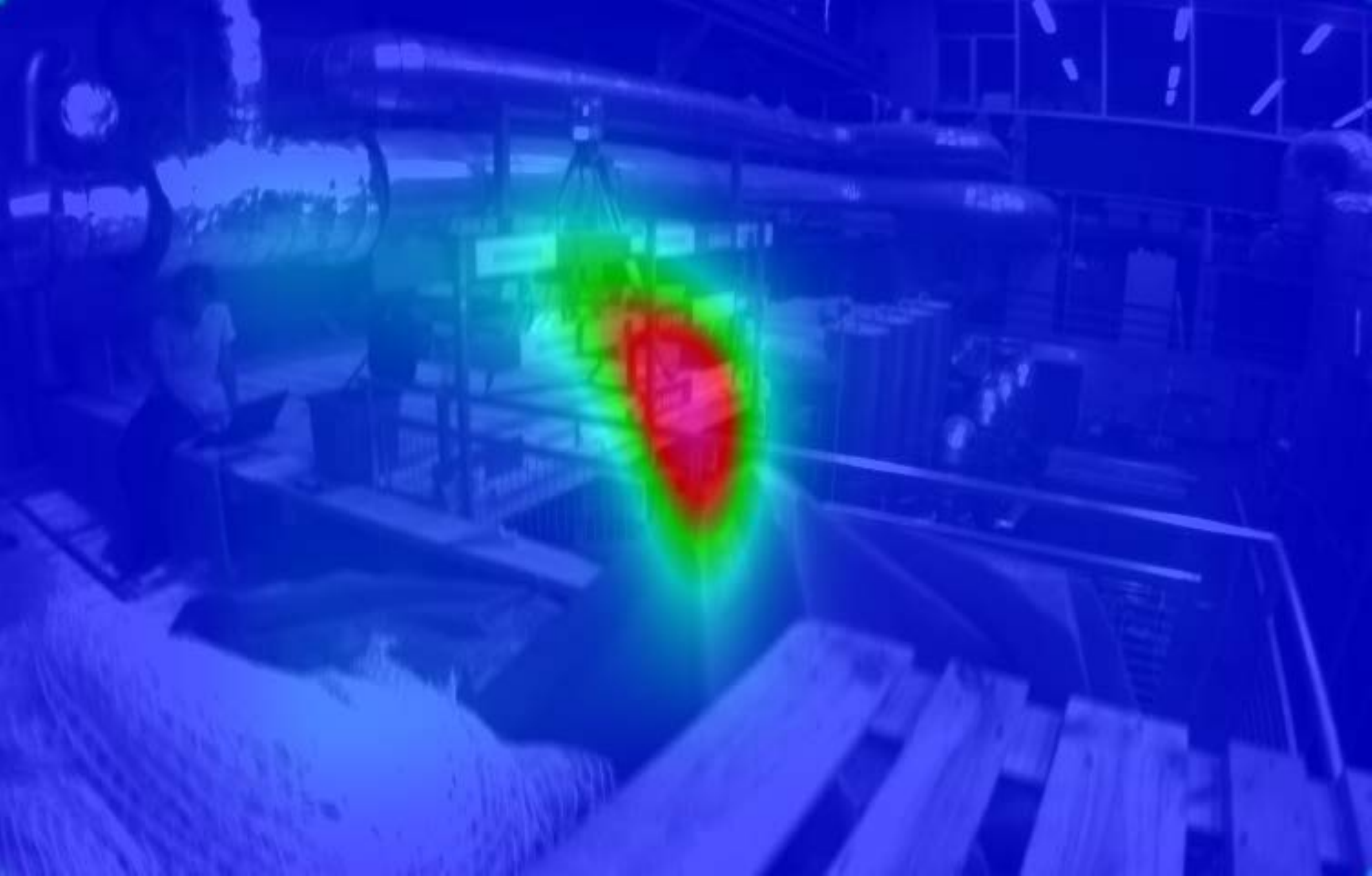}
		\end{minipage}%
		
	}%

	\centering
	\caption{Salient Euroc, a saliency SLAM datasets: (a) shows three consecutive images(from top to bottom), (b) is the corresponding saliency masks where the white parts indicate higher attention. In other words, these parts are more significant. In (c), we also combine the original images and the saliency masks into thermal maps for better visualization.}
	\label{fig: Salient Euroc}
\end{figure}

The Salient Euroc dataset includes the data of cam0 in the original dataset, the true value, and the corresponding saliency map. Fig. \ref{fig: Salient Euroc} shows three consecutive picture frames and their corresponding visual saliency masks in the salient Euroc dataset. It can be found that attention changes with the camera movement, but attention to salient objects are persistent.

\subsection{Comparison with Other SLAM Methods}
\subsubsection{Efficiency}
First, we compare the efficiency of SLAM to that of baseline methods. Increasing the effectiveness of the SLAM system will improve the applicability of the system. We have looked at the tracking time of the entire SLAM system. Mean Tracking Time is defined by equation (\ref{equation: mean tracking time}).
\begin{equation}
T_{mean} =  \dfrac{T_{total}}{N}
\label{equation: mean tracking time}
\end{equation}
where $ T_{mean} $ is the mean time to track every frame. $ T_{total} $ is the total running time of the SLAM system. $ N $ is the number of frames in every dataset.

Tab. \ref{table:Performance of efficiency} records the average runtime per frame on 11 Euroc datasets labeled with saliency maps. Our approach is more efficient than the baseline method on simple, medium, and difficult types of datasets. To more clearly analyze how our method improves efficiency, we define the following performance of efficiency:
\begin{equation}
P_{i} =  \dfrac{T_{Baseline}^i - T_{OurMethod}^i}{T_{Baseline}^i}
\end{equation}
where $ T_{Baseline}^i $ is the mean tracking time of the baseline method on the $i$-th dataset, and $ T_{OurMethod}^i $ is the mean tracking time of our approach on the $i$-th dataset
As shown in Tab. \ref{table:Performance of efficiency}, our method improves efficiency by an average of 8.43\%. It is because SalNavNet gives higher weight to the salient area of the images, and the lower weight to the insignificant area. When calculating in the SLAM system, the reprojection error of some points also becomes close to 0, thereby reducing the amount of calculation. The efficiency of MH01, MH03, MH05, V101, V103, V202 datasets has been enhanced by more than 10\%, but the performance of MH02 is the same as that of the baseline method. It is probably that in MH02 dataset, the feature points extracted by the baseline method mainly concentrated in the significant areas predicted by the saliency model. Hence, the efficiency of our approach is close to the baseline method.
\begin{table}[]
	\centering
	\caption{Comparison of efficiency between ORB-SLAM and Attention-SLAM. (Unit: seconds)}
	\label{table:Performance of efficiency}
	\begin{tabular}{ccc}
		\hline
		Seq.Name & ORB-SLAM  & Attention-SLAM     \\ \hline
		MH01     & 0.028305  & \textbf{0.0245115} \\
		MH02     & 0.0263993 & \textbf{0.0262359} \\
		MH03     & 0.025563  & \textbf{0.0227666} \\
		MH04     & 0.021658  & \textbf{0.020918}  \\
		MH05     & 0.0239042 & \textbf{0.0210823} \\
		V101     & 0.027441  & \textbf{0.0265265} \\
		V102     & 0.0236729 & 0.0243952          \\
		V103     & 0.0235398 & \textbf{0.0205895} \\
		V201     & 0.0245865 & 0.0245938          \\
		V202     & 0.0254648 & \textbf{0.0226415} \\
		V203     & 0.0219403 & \textbf{0.0203492} \\ \hline
	\end{tabular}
\end{table}

\subsubsection{Accuracy}
When evaluating the performance of the SLAM system, a common practice is to use Absolute Trajectory Error (ATE) to evaluate the performance of the SLAM system. It measures the difference between ground-truth and the trajectory estimated by the SLAM system. Firstly, we need to associate the estimated poses with ground-truth. Secondly, the trajectory generated by the monocular visual SLAM system does not have a scale, so we need to align the true and the estimated trajectory using singular value. Finally, we can compute the difference between each pair of poses and output the ATE. 

\begin{figure}[hbtp]
	\centering
	\subfloat[Trajectory of ORB-SLAM (V101)]{%
		\includegraphics[width=0.9\linewidth, height=0.9\linewidth]{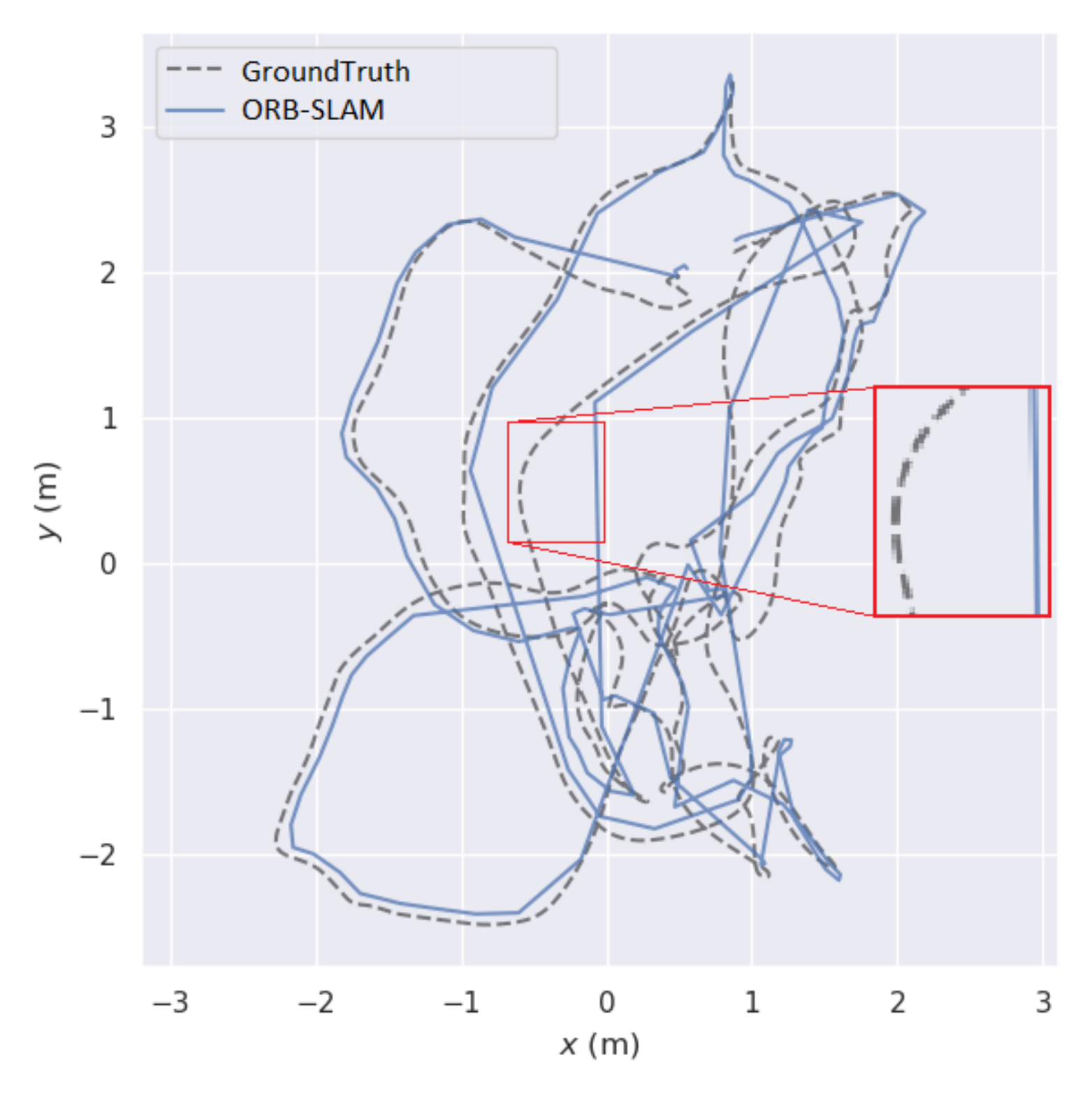}
		
	}
	
	\subfloat[Trajectory of Attention-SLAM (V101)]{
		\includegraphics[width=0.9\linewidth, height=0.9\linewidth]{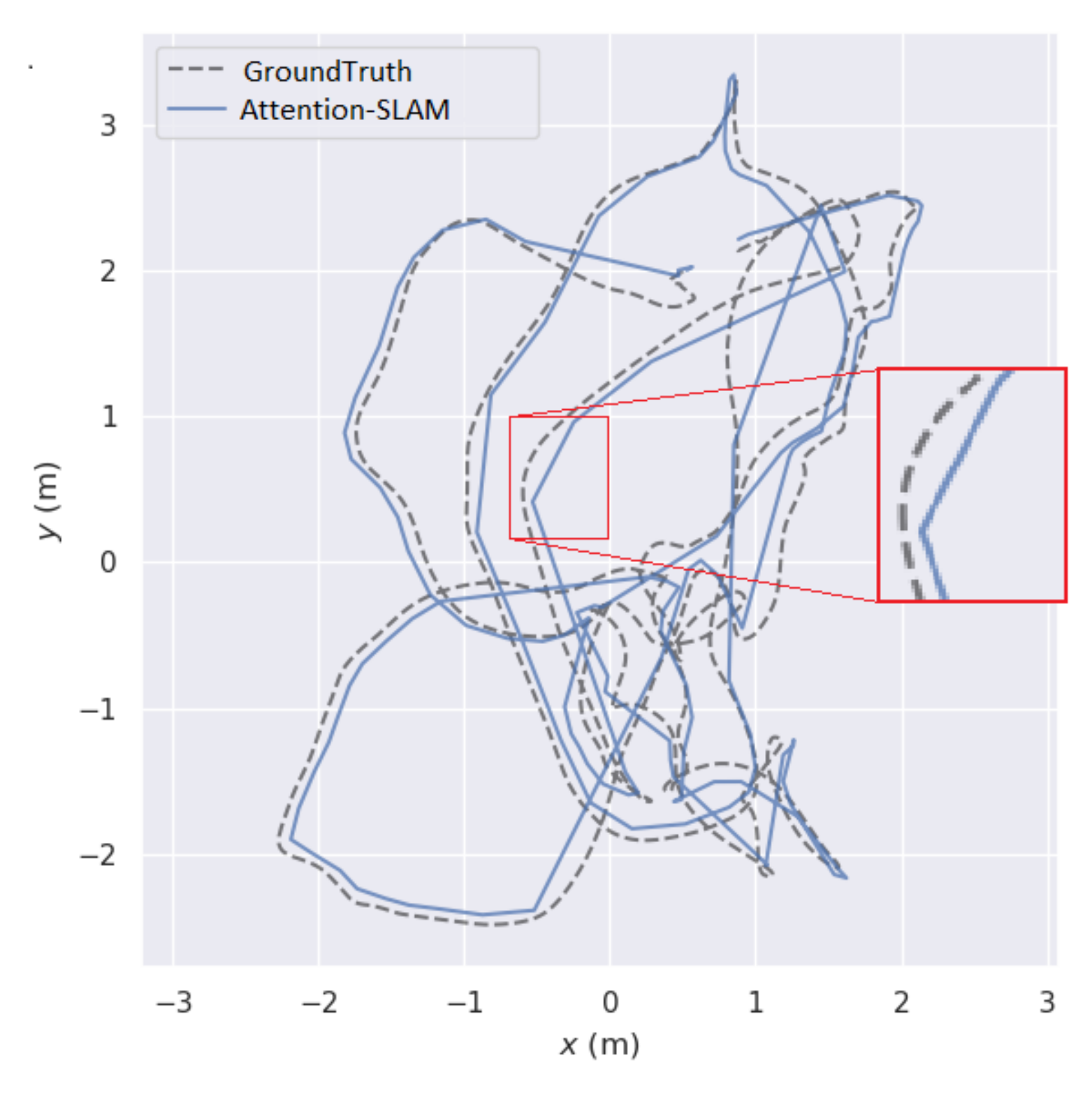}

	}
	
	\caption{The comparison of 2D trajectory between ORB-SLAM and Attention-SLAM on V101.}      
	\label{fig:XY Trajectory Comparison on V101}
\end{figure}
Fig. \ref{fig:XY Trajectory Comparison on V101} shows the 2D trajectory of our method and ORB-SLAM on the V101 dataset. It shows that the trajectory estimated by our method is much closer to the ground-truth. Our method pays more attention to salient feature points, thereby making the pose estimation closer to the true value. 
To better analyze the accuracy of our pose estimation, we plot the estimates of the 3D pose and the true value in Fig. \ref{fig: Trajectory Comparison}, respectively. We use a red frame to enlarge the significant parts of trajectories. Both methods track the trajectory well in the first 40 seconds, but after that, the baseline method has a large offset on the X-axis and Z-axis. At 50-60 seconds, our method tracks the Z-axis better than the baseline method. After 50 seconds of the Z-axis, the camera move up and down a lot, our method also has a smaller error than the baseline method.

To further evaluate our method, we compare the performance of Attention-SLAM with DSO \cite{33}, Salient DSO \cite{34} and ORB-SLAM \cite{7} in Tab. \ref{table: RmseATE} and Tab. \ref{table: meanATE}.  We calculate the mean ATE and RMSE ATE of our method and baseline methods. The results are as follows:
\begin{figure}[!hbtp]
	\centering
	\subfloat[Trajectory of ORB-SLAM (MH01)]{%
		\includegraphics[width=0.9\linewidth]{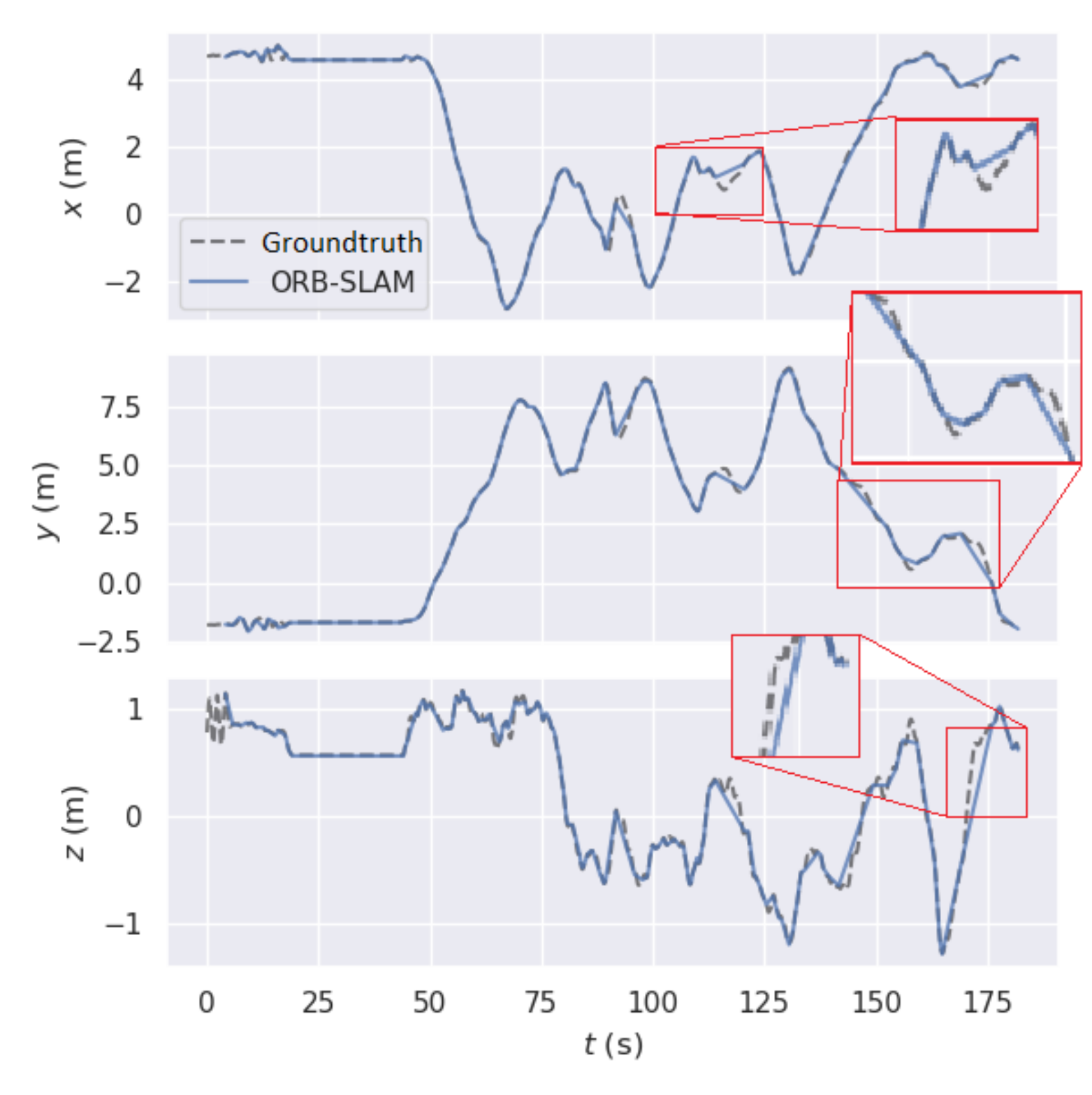}
	}
	
	\subfloat[Trajectory of Attention-SLAM (MH01)]{
		\includegraphics[width=0.9\linewidth]{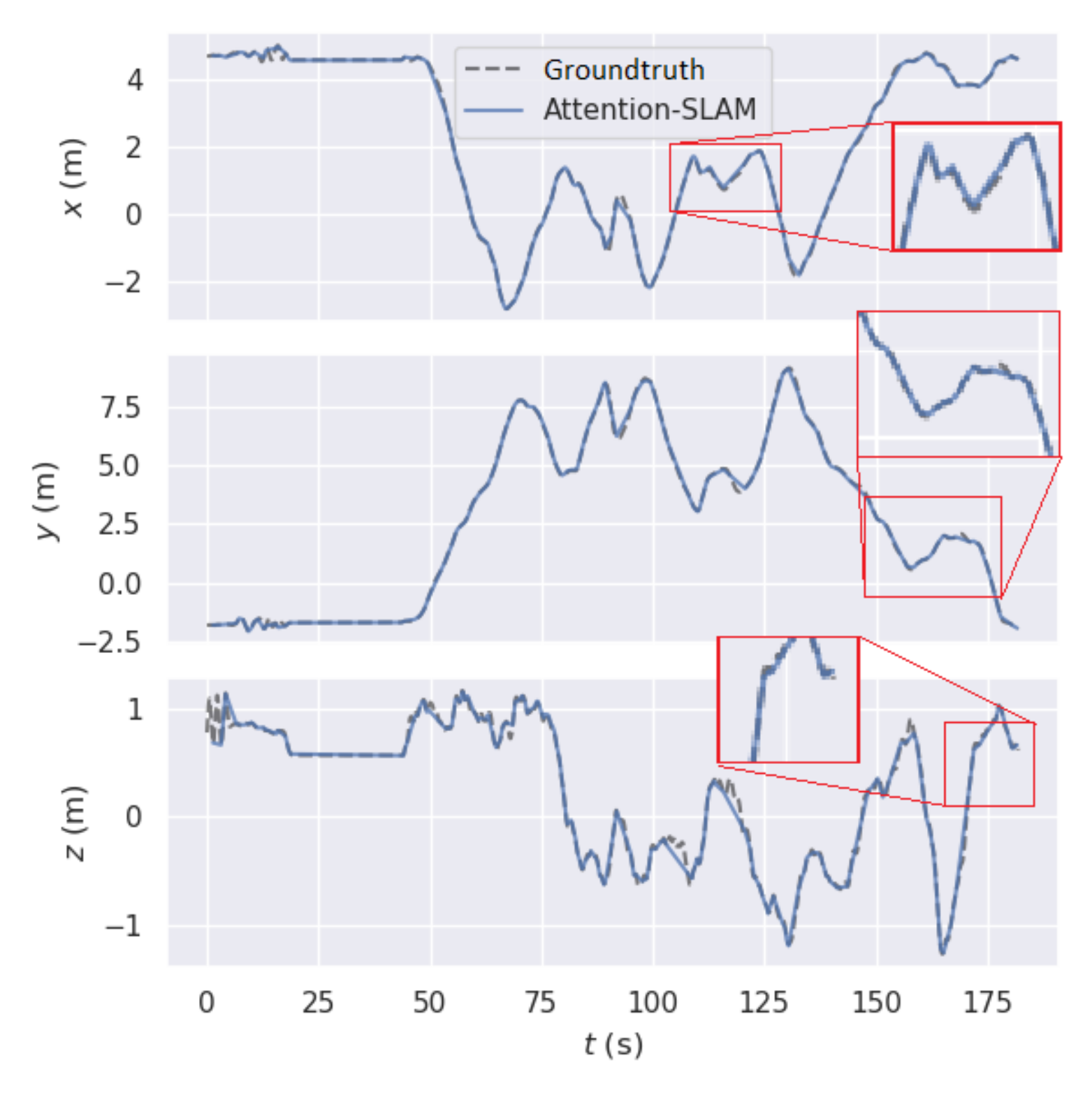}		
	}
	
	\caption{The 3D trajectory comparison of ORB-SLAM and Attention-SLAM.}
	\label{fig: Trajectory Comparison}
\end{figure}
\begin{table}[]
	\centering
	\caption{Mean Absolute Trajectory Error of related methods and Attention SLAM. (Unit: meter)}
	\label{table: meanATE}
	\begin{tabular}{ccccc}
		\hline
		Seq.Name         & DSO    & Salient DSO        & ORB-SLAM      & Attention-SLAM \\
		\hline
		MH01             & 0.0373 &\textbf{0.0344}     & 0.0384        & 0.0396\\
		MH02             & 0.0343 &0.0355              & 0.0299        &\textbf{0.0285}\\
		MH03             & 0.1687 &0.1313              &0.0349         & \textbf{0.0346}\\
		MH04             & 0.1531 &3.8666              &0.1051         & \textbf{0.0513}\\
		MH05             & 0.1056 &0.0887              &\textbf{0.0430}& 0.0445\\ \hline
		V101             & 0.1890 &0.0898              & 0.0926        & \textbf{0.0622}\\
		V102             & 0.3120 &\textbf{0.4628}     & 0.0625        & 0.0614 \\
		V103             & 0.7390 &0.7102              &\textbf{0.0655}& 0.0764\\ \hline
		V201             & 0.0764 &0.0689              & 0.05630       & \textbf{0.0534}\\
		V202             & 0.0822 &0.1103              & 0.0539        & \textbf{0.0532}\\
		V203             & 1.3712 &1.0069              &\textbf{0.0759}& 0.1043      \\ \hline
	\end{tabular}
\end{table}
\begin{table}[]
	\centering
	\caption{RMSE Absolute Trajectory Error of related methods and Attention-SLAM. (Unit: meter)}
	\label{table: RmseATE}
	\begin{tabular}{ccccc}
		\hline
		Seq.Name & DSO & Salient DSO & ORB-SLAM & Attention-SLAM \\
		\hline
		MH01& 0.0458& \textbf{0.0412}& 0.0449& 0.0449 \\
		MH02& 0.0423& 0.0435& 0.0346& \textbf{0.0328} \\
		MH03& 0.1937& 0.1522& 0.0387& \textbf{0.0371}          \\
		MH04& 0.1721& 4.5629& 0.1133& \textbf{0.0573} \\
		MH05& 0.1153& 0.0951& \textbf{0.0468}& 0.0472          \\ \hline
		V101& 0.2064& 0.0963& 0.0956 & \textbf{0.0951} \\
		V102& 0.6044& 0.6774& 0.0647 & \textbf{0.0631} \\
		V103& 0.8178& 0.7562& \textbf{0.0694}& 0.1020          \\ \hline
		V201& 0.0785& 0.0709& 0.0584& \textbf{0.0578} \\
		V202& 0.0938& 0.1322& 0.0569& \textbf{0.0565} \\
		V203& 1.5030& 1.2380&\textbf{0.0822} & 0.1161          \\ \hline
	\end{tabular}
\end{table}

From Tab. \ref{table: RmseATE} and Tab. \ref{table: meanATE}, Sailent DSO performs better than DSO, and our method performs better than ORB-SLAM. It convinces us that combining the saliency model with traditional monocular visual SLAM can enhance the performance of the conventional SLAM method. 

Results in Tab. \ref{table: RmseATE} and Tab. \ref{table: meanATE} shows that Attention-SLAM performs differently on different data sequences. Comparing Fig. \ref{table: RmseATE} with Tab. \ref{table: Classification of Datasets}, our method has higher accuracy in most cases. Among the same scenario, the accuracy of our approach decreased with the increase in difficulty. It shows that our approach on MH05, V103, and V203 dataset does not perform better than the baseline method. From the different characteristics of the dataset described in Tab. \ref{table:Euroc}, MH05, V103, and V203 have the characteristics of high-speed movement, which is the main reason for the decrease in tracking accuracy.

As illustrated above, we propose a saliency model to extract the essential information from the images sequence. Tab. \ref{table: KeyFrame} shows the keyframe number of ORB-SLAM and our method. 
\begin{table}[H]
	\caption{Number of keyframes}
	\label{table: KeyFrame}
	\centering
	\begin{tabular}{cccc}
		\hline
		Seq.Name             & ORB-SLAM\cite{7}     & Our Method         & Our Method + Entropy \\ \hline
		MH01                 & 204                  & 199                &\textbf{193}      \\
		MH02                 & 189                  & \textbf{186}       &\textbf{186}      \\
		MH03                 & 160                  & 164                &\textbf{155}         \\
		MH04                 & 199                  & \textbf{190}       &193      \\
		MH05                 & 223                  & \textbf{210}       &232      \\ \hline
		V101                 & 155                  & \textbf{153}       &\textbf{149}          \\
		V102                 & \textbf{152}         & 154                &165           \\
		V103                 & 195                  & \textbf{167}       &189          \\ \hline
		V201                 & \textbf{159}         & 160                &160          \\
		V202                 & \textbf{198}         & 200                &\textbf{196}       \\
		V203                 & 260                  & \textbf{256}       &266           \\ \hline
		\multicolumn{1}{l}{} & \multicolumn{1}{l}{} & \multicolumn{1}{l}{}
	\end{tabular}
\end{table}

Tab. \ref{table: Entropy Reduction} shows that our method has fewer keyframes on most dataset sequences but achieves higher accuracy. As described in equation (\ref{equation: entropy reduction}), we can get the entropy reduction between ORB-SLAM and Attention-SLAM on every data sequences in Tab. \ref{table: Entropy Reduction}. Our method reduces the uncertainty of the motion estimation on each frame. Besides, when we add the entropy ratio as an additional keyframe selection metric, it shows some interesting things. We compare the performance of Tab. \ref{table: meanATE} and Tab. \ref{table: KeyFrame}. Attention-SLAM performs well in the most easy and medium data sequences, but don't perform well in difficult sequences, such as MH04, MH05, V203, V103. After we add the entropy keyframe selection criteria to Attention-SLAM, this criteria makes Attention-SLAM select more keyframes in difficult data sequences. Results in Tab. \ref{table: Comparison between at} show that this criteria makes Attention-SLAM perform better in difficult data sequences. So the entropy ratio metric is an excellent implement to our method. When the saliency model adds enough semantic information to the system, it makes the system choose fewer keyframes. When the saliency model unable to reduce the uncertainty of motion estimation, it makes the system select more keyframes to achieve better performance.
\begin{table}[H]
	\caption{Mean ATE Performance comparison of Attention-SLAM before and after adding entropy ratio selection. (Unit: meter)}
	\label{table: Comparison between at}
	\centering
	\begin{tabular}{ccc}
		\hline
		Seq.Name & Attention-SLAM & Attention-SLAM + Entropy Selection \\ \hline
		MH01     & \textbf{0.3637}         & 0.3767                                    \\
		MH02     & 0.2849         & \textbf{0.2715}                                    \\
		MH03     & 0.0439         & \textbf{0.0351}                                    \\
		MH04     & \textbf{0.0668}         & 0.130                                     \\
		MH05     & 0.0455         & \textbf{0.0413}                                    \\ \hline
		V101     & \textbf{0.0917}         & 0.0924                                    \\
		V102     & 0.06331        & \textbf{0.06239}                                   \\
		V103     & 0.0668         & \textbf{0.0650}                                   \\ \hline
		V201     & \textbf{0.0535}         & 0.0546                                    \\
		V202     & \textbf{0.053}          & 0.0568                                    \\
		V203     & 0.10956        & \textbf{0.08895}                                   \\ \hline
	\end{tabular}
\end{table}
\begin{table}[H]
	\caption{Entropy reduction of our method}
	\label{table: Entropy Reduction}
	\centering
	\begin{tabular}{cccc}
		\hline
		Seq.Name & ORB-SLAM & Our Method & Entropy Reduction(bit) \\ \hline
		MH01     & 26.0373  & 19.4524    & 0.1266            \\
		MH02     & 22.8747  & 19.0783    & 0.0788            \\
		MH03     & 23.8194  & 19.8594    & 0.0790            \\
		MH04     & 17.3203  & 12.3276    & 0.1476            \\
		MH05     & 19.0109  & 14.0694    & 0.1307            \\
		V101     & 25.8     & 24.6416    & 0.0199            \\
		V102     & 20.4733  & 15.7103    & 0.1150            \\
		V103     & 13.4433  & 11.396     & 0.0717            \\
		V201     & 24.6734  & 23.0403    & 0.0297            \\
		V202     & 13.7842  & 12.7896    & 0.0325            \\
		V203     & 7.70454  & 6.10752    & 0.1009            \\ \hline
	\end{tabular}
\end{table}
In Tab. \ref{table: Entropy Reduction} shows that our method reduces the uncertainty of traditional methods. And we can see that the entropy reduction has a positive correlation with the accuracy of our approach.
\subsubsection{Robustness}
To analyze the robustness of Attention-SLAM, we first compare the difficulty of the Euroc dataset with the results of the previous section.\\
\begin{table}[H]
	\caption{Classification of datasets}
	\label{table: Classification of Datasets}
	\centering
	\begin{tabular}{c c}
		\hline
		Difficulty & Dataset \\ 
		\hline
		Easy & MH01, MH02, V101, V201           \\
		Medium & MH03, V102, V202               \\
		Difficulty & MH04, MH05, V103, V203     \\        
		\hline
	\end{tabular}
\end{table}
As shown in Tab. \ref{table: RmseATE}, Tab. \ref{table: meanATE}, Attention-SLAM improves the accuracy on most datasets but doesn't perform well on datasets that contain motion blur and fast rotation. It may lead our saliency model unable to find the correct significant area in the frames. We will discuss this more with details in the Discussion Section.

\section{Discussion}
In the experiment, the saliency weight generated by the image saliency model SalGAN made Attention-SLAM achieve the best performance on SLAM metrics. In the video saliency model, we introduced the correlation module and proposed an adaptive EMA module. The SalNavNet help Attention-SLAM have higher accuracy. In theory, SalEMA helped Attention-SLAM should perform better than what SalGAN did because the video saliency model can use LSTM to learn the continuous information of adjacent frames. However, as the ATE results shown in Tab. \ref{table: Comparison between different saliency models}, the saliency maps generated by SalGAN enabled Attention-SLAM to achieve the best performance. This contrast between theory and reality has caught our attention. We found that this is mainly due to the different distribution of the datasets for training the two models. The training dataset of SalGAN is SALICON, while SalEMA was mainly trained on the DHF1K dataset. From the distribution of the true saliency value of the dataset, SALICON has a much weaker center bias than DHF1K, which can potentially be explained by the longer viewing time of each image/frame (5s vs. 30ms to 42ms) that allows secondary stimuli to be fixated. 

Therefore, the saliency map generated by the SalGAN model also showed a much weaker center bias. SalGAN can continuously maintain attention to the same salient object. This made the saliency map generated by SalGAN help the SLAM system get higher accuracy.

The dataset used to train the video saliency model, such as DHF1K has a relatively strong center bias. The ground-truth saliency value of this dataset is mostly distributed in the middle of the video, making the saliency maps generated by SalEMA mostly distributed in the middle of the image. When a salient object moved from the center of the image to the edge of the image, saliency maps of SalEMA choose to refocus on other objects located in the center of the image. This phenomenon made the saliency map generated by SalEMA unable to help the SLAM system continuously track the same salient key points. On one hand, SalNavNet enhanced the ability of the saliency model to learn adjacent frame information. On the other hand, it mitigated the center bias problem caused by the data distribution of DHF1K.  When objects moved to the edge of the image with the camera, the saliency map generated by SalNavnet can track them continuously.

We found that the motion blur caused by high-speed motion in the Euroc dataset affects the accuracy of SLAM system. The motion blur makes the SLAM system unable to extract enough feature points, resulting in a temporary tracking loss in the SLAM system.
\section{Conclusion}
This paper proposed a semantic SLAM method called Attention-SLAM. It combined visual saliency semantic information and visual SLAM system. SalNavNet is designed for the purpose of navigation is also proposed in this paper. We established a dataset based on the Euroc dataset namely Salient Euroc, a SLAM dataset labeled with saliency semantic information. Compared with the current mainstream monocular visual SLAM methods, our method has higher efficiency and accuracy. We also analyzed our method from the perspective of information theory. The results proved that visual saliency information could reduce the uncertainty of pose estimation.

As a future research direction, it is necessary to establish a specialized visual saliency dataset for navigation. The current saliency datasets have a strong center bias, this is the essential reason why the existing saliency models cannot help Attention-SLAM achieve higher accuracy. If we build a saliency navigation dataset, it will solve the problem of unable to focus on the same object continuously and improve the accuracy of our method further. Morever, we will tightly couple the visual saliency model into the vision SLAM system to further reduce the uncertainty and improve the efficiency in the futune study.
 \section*{Acknowledgement}
This work was supported by the National Nature Science Foundation of China (NSFC) under Grant 61873163, Equipment Pre-Research Field Foundation under Grant 61405180205. 

We want to express our gratitude to Dr. Lei Chu from Shanghai Jiao Tong University for paper writing and building the visual saliency model.
\appendices

\ifCLASSOPTIONcaptionsoff
  \newpage
\fi


\begin{IEEEbiography}[{\includegraphics[width=1in,height=1.25in,clip,keepaspectratio]{lia}}]{Jinquan Li}
    received the B.S degree in electronic engineering from Xidian University, Xian, China in 2019. He is currently pursing the M.Sc degree in Shanghai Jiao Tong University, China.
    His current research interests include semantic SLAM and visual saliency.
\end{IEEEbiography}
\begin{IEEEbiography}[{\includegraphics[width=1in,height=1.25in,clip,keepaspectratio]{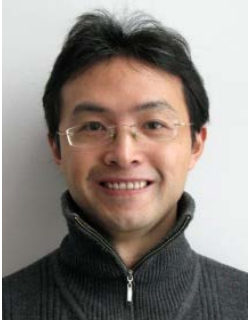}}]{Ling Pei}
	received the Ph.D. degree from Southeast University, Nanjing, China, in 2007.From 2007 to 2013, he was a Specialist Research Scientist with the Finnish Geospatial Research Institute. He is currently an Associate Professor with the School of Electronic Information and Electrical Engineering, Shanghai Jiao Tong University. He has authored or co-authored over 90 scientific papers. He is also an inventor of 24 patents and pending patents. His main research is in the areas of indoor/outdoor seamless positioning, ubiquitous computing, wireless positioning, Bio-inspired navigation, context-aware applications, location-based services, and navigation of unmanned systems. Dr. Pei was a recipient of the Shanghai Pujiang Talent in 2014.
\end{IEEEbiography}
\begin{IEEEbiography}[{\includegraphics[width=1in,height=1.25in,clip,keepaspectratio]{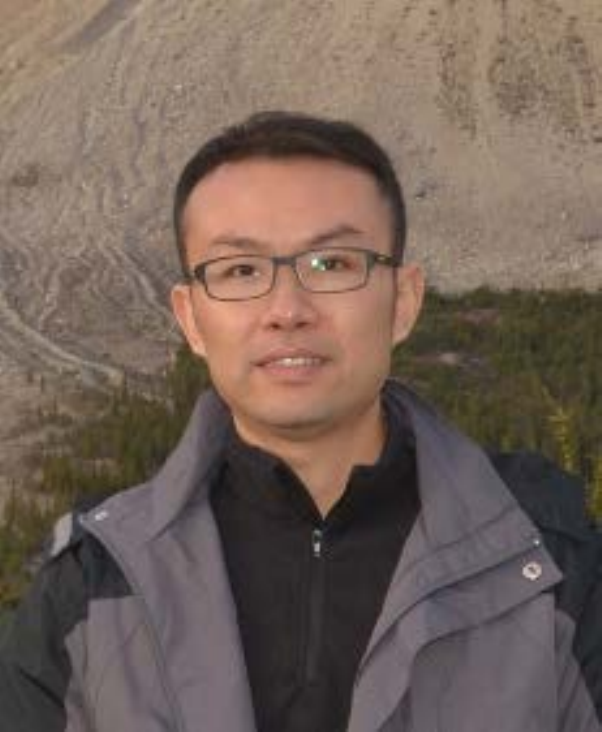}}]{Danping Zou}
	received the B.S. degree from Huazhong University of Science and Technology, Wuhan, China, and the Ph.D. degree from Fudan University, Shanghai, China, in 2003 and 2010, respectively. From 2010 to 2013, he was a Research Fellow with the Department of Electrical and Computer Engineering, National University of Singapore, Singapore. In 2013, he joined the Department of Electronic Engineering, Shanghai Jiao Tong University, Shanghai, as an Associate Professor. His research interests include low-level 3-D vision on robotics.
\end{IEEEbiography}
\begin{IEEEbiography}[{\includegraphics[width=1in,height=1.25in,clip,keepaspectratio]{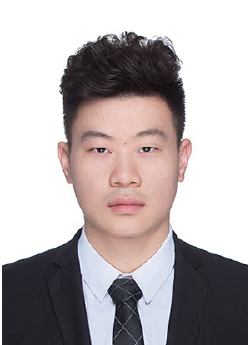}}]{Songpengcheng Xia}
	received the B.S. degree in navigation engineering from Wuhan University, Wuhan, China, in 2019. He is currently pursuing the Ph.D. degree with Shanghai Jiao Tong University, Shanghai, China. \\
	His current research interests include machine learning, inertial navigation systems, multi-sensor fusion and mixed-reality simulation technology
\end{IEEEbiography}
\begin{IEEEbiography}[{\includegraphics[width=1in,height=1.25in,clip,keepaspectratio]{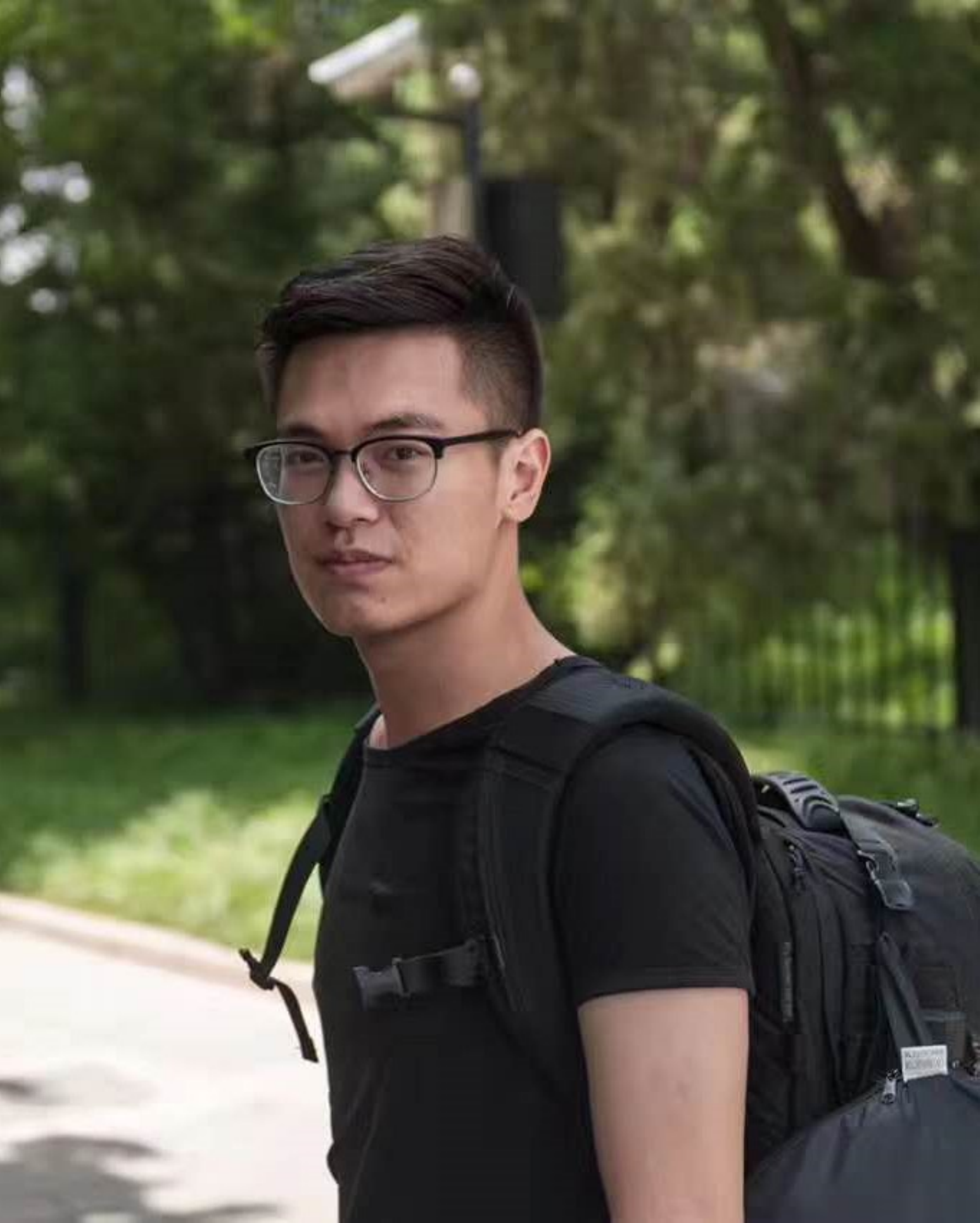}}]{Qi Wu}
	received the B.S. degree in Chongqing University of Posts and Telecommunications, Chongqing, China, in 2018.He received the M.S. degree in Beijing University of Posts and Telecommunications, Beijing, China. He is currently working toward the
	Ph.D degree in Shanghai Jiao Tong University. His main research interests include visual-SLAM ,Lidar-SLAM, Multi-sensor fusion.
\end{IEEEbiography}
\begin{IEEEbiography}[{\includegraphics[width=1in,height=1.25in,clip,keepaspectratio]{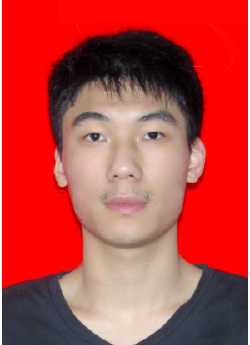}}]{Tao Li}
	received the B.S. degree in navigation engineering from Wuhan University, Wuhan, China, in 2018. 
	He is currently pursuing the Ph.D. degree with Shanghai Jiao Tong University, Shanghai, China. 
	His current research interests include visual simultaneous localization and mapping (SLAM), 
	light detection and ranging (LiDAR)-SLAM, global navigation satellite systems(GNSS), 
	inertial navigation systems (INS), and information fusion.
\end{IEEEbiography}
\begin{IEEEbiography}[{\includegraphics[width=1in,height=1.25in,clip,keepaspectratio]{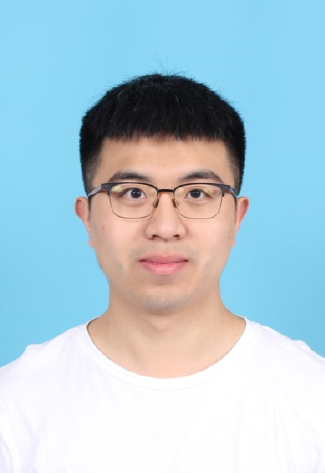}}]{Zhen Sun}
	Zhen Sun received the M.S. degree in Shanghai Jiaotong University, Shanghai, China, in 2020. He is currently working toward the Ph.D degree in SJTU. His main research direction is cognitive navigation.
\end{IEEEbiography}
\begin{IEEEbiography}[{\includegraphics[width=1in,height=1.25in,clip,keepaspectratio]{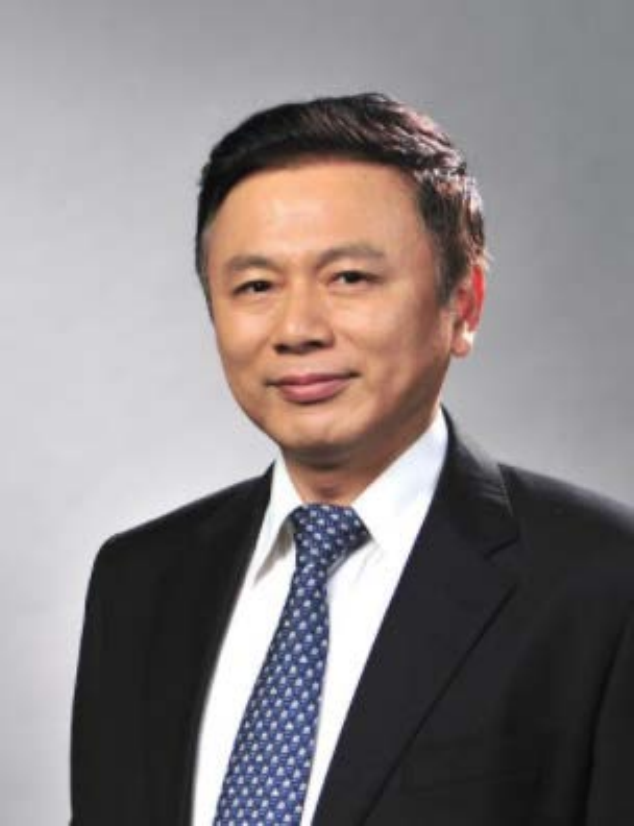}}]{Wenxian Yu}
	received the B.S., M.S., and Ph.D. degrees from the National University of Defense Technology, Changsha, China, 
	in 1985, 1988, and 1993, respectively. From 1996 to 2008, he was a Professor with the College of Electronic Science 
	and Engineering, National University of Defense Technology, where he was also the Deputy Head of the 
	College and an Assistant Director of the National Key Laboratory of Automatic Target Recognition. 
	From 2009 to 2011, he was the Executive Dean of the School of Electronic, Information, and Electrical Engineering, 
	Shanghai Jiao Tong University, Shanghai, China. He is currently a Yangtze River Scholar Distinguished Professor 
	and the Head of the research part in the School of Electronic, Information, and Electrical Engineering, Shanghai Jiao Tong University. 
	His research interests include remote sensing information processing, automatic target recognition, multisensor data fusion, etc.
\end{IEEEbiography}

\end{document}